\pdfoutput=1
\documentclass[11pt]{article}

\usepackage{acl}

\usepackage{times}
\usepackage{latexsym}

\usepackage[T1]{fontenc}
\usepackage[utf8]{inputenc}

\usepackage{tikz}
\usepackage{subfig}
\usepackage{xcolor}
\usepackage{tcolorbox}

\usepackage{multirow}
\usepackage{booktabs}
\usepackage{pgfplots}
\usepackage{amssymb}
\usepackage{array}
\usepackage{xspace}
\usepackage{pgfplotstable}
\usepackage{pgf}

\usetikzlibrary{plotmarks}
\usetikzlibrary{backgrounds}
\usetikzlibrary{fit}
\usetikzlibrary{decorations}
\usetikzlibrary{decorations.markings}
\usetikzlibrary{positioning}

\definecolor{myblue}{RGB}{31,120,180}
\definecolor{mygreen}{RGB}{51,160,44}
\definecolor{myred}{RGB}{227,26,28}

\definecolor{iyellow}{RGB}{255,250,205}
\definecolor{ipurple}{RGB}{230,230,250}

\definecolor{upurple}{RGB}{155,89,182}
\definecolor{ublue}{RGB}{52,152,219}
\definecolor{ured}{RGB}{231,76,60}
\definecolor{udark}{RGB}{77,153,77}
\definecolor{ugreen}{RGB}{46,204,113}

\usepackage{microtype}

\title{ODE Transformer: An Ordinary Differential Equation-Inspired Model for Sequence Generation}

\author{
  Bei Li$^1$,
  Quan Du$^{1,2}$,
  Tao Zhou$^1$,
  Yi Jing$^1$, Shuhan Zhou$^1$, Xin Zeng$^1$ \\
  \textbf{Tong Xiao$^{1,2}$\thanks{\xspace\xspace Corresponding author.}},
  \textbf{Jingbo Zhu$^{1,2}$},
  \textbf{Xuebo Liu$^{3}$}
  \textbf{and Min Zhang$^{3}$} \\
  $^{1}$School of Computer Science and Engineering, Northeastern University, Shenyang, China\\
  $^{2}$NiuTrans Research, Shenyang, China \\
  $^{3}$Institute of Computing and Intelligence, Harbin Institute of Technology, Shenzhen, China\\
  {\tt
        libei\_neu@outlook.com, \{xiaotong,zhujingbo\}@mail.neu.edu.cn,
  }\\
  {\tt
        \{liuxuebo,zhangmin2021\}@hit.edu.cn
  }
}

\begin{document}
\maketitle
\begin{abstract}
  Residual networks are an Euler discretization of solutions to Ordinary Differential Equations (ODE). This paper explores a deeper relationship between Transformer and numerical ODE methods. We first show that a residual block of layers in Transformer can be described as a higher-order solution to ODE. Inspired by this, we design a new architecture, {\it ODE Transformer}, which is analogous to the Runge-Kutta method that is well motivated in ODE. As a natural extension to Transformer, ODE Transformer is easy to implement and efficient to use. Experimental results on the large-scale machine translation, abstractive summarization, and grammar error correction tasks demonstrate the high genericity of ODE Transformer. It can gain large improvements in model performance over strong baselines (e.g., 30.77 and 44.11 BLEU scores on the WMT'14 English-German and English-French benchmarks) at a slight cost in inference efficiency.
  % It has been found that residual networks are an Euler discretization of solutions to Ordinary Differential Equations (ODEs). In this paper, we explore a deeper relationship between Transformer and numerical methods of ODEs. We show that a residual block of layers in Transformer can be described as a higher-order solution to ODEs. This leads us to design a new architecture (call it ODE Transformer) analogous to the Runge-Kutta method that is well motivated in ODEs. As a natural extension to Transformer, ODE Transformer is easy to implement and parameter efficient. Our experiments on large-scale machine translation, abstractive summarization and grammar error correction tasks demonstrate the genericity of this model, and large improvements in performance over several strong baselines. It achieves 30.77 and 44.11 BLEU scores on the WMT'14 En-De and En-Fr test data.

  % This sets a new state-of-the-art on the WMT'14 En-Fr task without leveraging data augmentation or pre-training techniques.
\end{abstract}

\section{Introduction}

Residual networks have been used with a great success as a standard method of easing information flow in multi-layer neural models~\cite{he2016deep,vaswani2017attention}. Given an input $y_t$, models of this kind define the output of a layer $t$ to be:
\begin{eqnarray}
y_{t+1} & = & y_t + F(y_t, \theta_t) \label{eq:residual}
\end{eqnarray}
\noindent where $F(\cdot,\cdot)$ is the function of the layer and $\theta_t$ is its parameter. Interestingly, recent work in machine learning \cite{weinan2017proposal,yiping2018beyond,haber2018learning,chang2018reversible,ruthottohaber2019} points out that Eq. (\ref{eq:residual}) is an Euler discretization of the Ordinary Differential Equation (ODE), like this:
\begin{eqnarray}
\frac{\mathrm{d} y(t)}{\mathrm{d} t} & = & F(y(t), \theta(t)) \label{eq:ode}
\end{eqnarray}
\noindent where $y(t)$ and $\theta(t)$ are continuous with respect to $t$. In this way, we can call Eq. (\ref{eq:residual}) an \textit{ODE block}. This finding offers a new way of explaining residual networks in the view of numerical algorithms. Then, one can think of a multi-layer network as applying the Euler method (i.e., Eq. (\ref{eq:residual})) to solve Eq. (\ref{eq:ode}) subject to the initial conditions $y(0)=y_0$ and $\theta(0)=\theta_0$.

% ##########Figure 1############
%----------------------------------------------
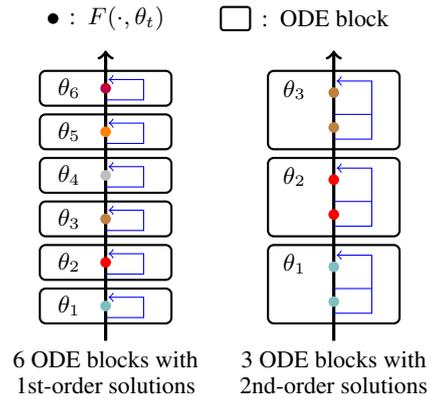
\begin{figure}[t!]
    \centering
    \begin{tikzpicture}
  
    \newlength{\seg}
    \setlength{\seg}{1.5em}
    \newlength{\subseg}
    \setlength{\subseg}{8em}
    \newlength{\resseg}
    \setlength{\resseg}{1.3em}
  
    \tikzstyle{circlenode}=[circle,minimum size=4pt,inner sep=0,fill=blue!80];
    \tikzstyle{odenode}=[draw,minimum height=1.2em,minimum width=4.5em,inner sep=0,thick,rounded corners=.2em];

    % sub figure (standard)
    \draw [->,very thick] (0,0) -- (0,10em);
    \node [anchor=north,font=\footnotesize,align=center] (cap01) at (0,0) {6 ODE blocks with \\ 1st-order solutions };
    %\node [anchor=north west] (cap02) at ([yshift=0.3em]cap01.south west) {\footnotesize{1st-order solutions}};
  
    \node [circlenode,fill=teal!50] (node01) at (0,-0.3em+1*\seg) {};
    \node [circlenode,fill=red] (node02) at (0,-0.3em+2*\seg) {};
    \node [circlenode,fill=brown] (node03) at (0,-0.3em+3*\seg) {};
    \node [circlenode,fill=gray!50] (node04) at (0,-0.3em+4*\seg) {};
    \node [circlenode,fill=orange] (node05) at (0,-0.3em+5*\seg) {};
    \node [circlenode,fill=purple] (node06) at (0,-0.3em+6*\seg) {};
  
    \foreach \i in {1,...,6}
    {
        \node [odenode] (layer0\i) at (node0\i) {};
        \node [anchor=west] (label0\i) at ([xshift=0.3em]layer0\i.west) {\footnotesize{$\theta_\i$}};
        \draw [thin,->,blue] ([yshift=-0.4em]node0\i.center) -- ([yshift=-0.4em,xshift= \resseg]node0\i.center) -- ([yshift=0.3em,xshift= \resseg]node0\i.center) -- ([yshift=0.3em,xshift=0.1em]node0\i.center);
    }
  
    % sub figure (high-order ODE)
    \draw [->,very thick] (-0.2em+\subseg,0) -- (-0.2em+\subseg,10em);
    \node [anchor=north,font=\footnotesize,align=center] (cap11) at (-0.2em+\subseg,0) {3 ODE blocks with \\ 2nd-order solutions};
    %\node [anchor=north west] (cap12) at ([yshift=0.3em]cap11.south west) {\footnotesize{2rd-order solutions}};
  
    \node [circlenode,fill=teal!50] (node11) at (-0.2em+\subseg,-0.3em+1*\seg+0.1*\seg) {};
    \node [circlenode,fill=teal!50] (node12) at (-0.2em+\subseg,-0.3em+2*\seg-0.1*\seg) {};
    \node [circlenode,fill=red] (node13) at (-0.2em+\subseg,-0.3em+3*\seg+0.1*\seg) {};
    \node [circlenode,fill=red] (node14) at (-0.2em+\subseg,-0.3em+4*\seg-0.1*\seg) {};
    \node [circlenode,fill=brown] (node15) at (-0.2em+\subseg,-0.3em+5*\seg+0.1*\seg) {};
    \node [circlenode,fill=brown] (node16) at (-0.2em+\subseg,-0.3em+6*\seg-0.1*\seg) {};
  
    \node [odenode,minimum height=2.7em] (layer11) at ([yshift=0.4*\seg]node11) {};
    \node [odenode,minimum height=2.7em] (layer12) at ([yshift=0.4*\seg]node13) {};
    \node [odenode,minimum height=2.7em] (layer13) at ([yshift=0.4*\seg]node15) {};
    \node [anchor=north west] (label11) at ([xshift=0.2em]layer11.north west) {\footnotesize{$\theta_1$}};
    \node [anchor=north west] (label12) at ([xshift=0.2em]layer12.north west) {\footnotesize{$\theta_2$}};
    \node [anchor=north west] (label13) at ([xshift=0.2em]layer13.north west) {\footnotesize{$\theta_3$}};
    \draw [thin,->,blue] ([yshift=-0.4em]node11.center) -- ([yshift=-0.4em,xshift=\resseg]node11.center) -- ([yshift=0.4em,xshift=\resseg]node12.center) -- ([yshift=0.4em,xshift=0.1em]node12.center);
    \draw [thin,-,blue] ([yshift=0.3*\seg]node11.center)  -- ([yshift=0.3*\seg,xshift=\resseg]node11.center);
  
    \draw [thin,->,blue] ([yshift=-0.4em]node13.center) -- ([yshift=-0.4em,xshift=\resseg]node13.center) -- ([yshift=0.4em,xshift=\resseg]node14.center) -- ([yshift=0.4em,xshift=0.1em]node14.center);
    \draw [thin,-,blue] ([yshift=0.3*\seg]node13.center)  -- ([yshift=0.3*\seg,xshift=\resseg]node13.center);
  
    \draw [thin,->,blue] ([yshift=-0.4em]node15.center) -- ([yshift=-0.4em,xshift=\resseg]node15.center) -- ([yshift=0.4em,xshift=\resseg]node16.center) -- ([yshift=0.4em,xshift=0.1em]node16.center);
    \draw [thin,-,blue] ([yshift=0.3*\seg]node15.center)  -- ([yshift=0.3*\seg,xshift=\resseg]node15.center);
    %\draw [thin,-,blue] ([yshift=0.3*\seg]node12.center)  -- ([yshift=0.3*\seg,xshift=\resseg]node12.center);
    %\draw [thin,->,blue] ([yshift=-0.6em]node14.center) -- ([yshift=-0.6em,xshift= \resseg]node14.center) -- ([yshift=0.6em,xshift=\resseg]node16.center) -- ([yshift=0.6em,xshift=0.1em]node16.center);
    %\draw [thin,-,blue] ([yshift=0.3*\seg]node14.center)  -- ([yshift=0.3*\seg,xshift=\resseg]node14.center);
    %\draw [thin,-,blue] ([yshift=0.3*\seg]node15.center)  -- ([yshift=0.3*\seg,xshift=\resseg]node15.center);
  
    % lengend
    \node [anchor=west,circlenode,fill=black] (lengend01) at ([yshift=2.2em,xshift=-2em]node06.north) {};
    %\node [anchor=west,thick,draw,inner sep=0,minimum height=0.8em,minimum width=1em,rounded corners=0.1em] (tmp) at ([xshift=1em]layer13.east) {};
    \node [anchor=west,draw,thick,inner sep=0,minimum height=0.8em,minimum width=1em,rounded corners=0.1em] (lengend03) at ([xshift=5.5em]lengend01.east) {};
    \node [anchor=west] (lengend02) at ([xshift=0em]lengend01.east) {\small{:\ \ $F(\cdot,\theta_t)$}};
    \node [anchor=west] (lengend04) at ([xshift=0em]lengend03.east) {\small{:\ \ ODE block}};
  
    \end{tikzpicture}
    \caption{Models with different ODE blocks. }
    \label{fig:ode-blocks}
    \end{figure}  
  %----------------------------------------------
% ##########Figure 1############

The solution of Eq. (\ref{eq:ode}) has a sufficiently low error bound (call it a \textit{stable solution}) only if $\theta(t)$ changes slow along $t$ \cite{haberruthotto2017,chen2018neural}. But this assumption does not always hold for state-of-the-art natural language processing (NLP) systems, in which models are non-linear and over-parameterized. For example, language modeling and machine translation systems learn quite different parameters for different layers, especially when the layers are close to the model input \cite{vaswani2017attention,dai-etal-2019-transformer}. Also, truncation errors are nonnegligible for the Euler method because it is a first-order approximation to the true solution \cite{he2019ode}. These problems make the situation worse, when more layers are stacked and errors are propagated through the neural network. It might explain why recent Machine Translation (MT) systems cannot benefit from extremely deep models \cite{wang-etal-2019-learning,liu-etal-2020-understanding,wei-etal-2020-multiscale,li-etal-2020-shallow}.

This paper continues the line of research on the ODE-inspired method. The basic idea is to use a high-order method for more accurate numerical solutions to the ODE. This leads to a larger ODE block that generates a sequence of intermediate approximations to the solution. We find that the larger ODE block is sufficient to take the role of several ODE blocks with first-order solutions. The benefit is obvious: the use of fewer ODE blocks lowers the risk of introducing errors in block switching, and the high-order method reduces the approximation error in each ODE block. See Figure \ref{fig:ode-blocks} for a comparison of different models.

Our method is parameter-efficient because $\theta(t)$ is re-used within the same ODE block. As another ``bonus", the model can be improved by learning coefficients of different intermediate approximations in a block.
We evaluate our method in strong Transformer systems, covering both the wide (and big) model and the deep model.
For machine translation tasks, ODE Transformer achieves 30.77 and 44.11 BLEU scores on the WMT'14 En-De and En-Fr test sets, setting a new state-of-the-art on the WMT'14 En-Fr task.
It also significantly outperforms baselines on abstractive summarization and grammar error correction tasks.

\section{Transformer and ODEs}

We start with a description of Transformer, followed by its relationship with ODEs.
We choose Transformer for our discussion and experiments because it is one of the state-of-the-art models in recent sentence generation tasks.

\subsection{Transformer}
Transformer is an example of the encoder-decoder paradigm \cite{vaswani2017attention}. The encoder is a stack of identical layers. Each layer consists of a self-attention block and a feedforward network (FFN) block. Both of them equip with a residual connection and a layer normalization unit. Note that the term ``block'' is used in many different ways. In this paper, the term refers to any neural network that is enhanced by the residual connection (occasionally call it a \textit{residual block}). Following the Pre-norm architecture \cite{wang-etal-2019-learning}, we define a block as
\begin{eqnarray}
y_{t+1} & = & y_t + G(\textrm{LN}(y_t), \theta_t) \label{eq:prenorm}
\end{eqnarray}
\noindent where $\textrm{LN}(\cdot)$ is the layer normalization function,\footnote{We drop the parameter of $\textrm{LN}(\cdot)$ for simplicity.} and $G(\cdot)$ is either the self-attention or feedforward network. The decoder shares a similar architecture, having an additional encoder-decoder attention block sandwiched between the self-attention and FFN blocks.

\subsection{Ordinary Differential Equations}

An ordinary differential equation is an equation involving a function $y(t)$ of a variable $t$ and its derivatives. A simple form of ODE is an equation that defines the first-order derivative of $y(t)$, like
\begin{eqnarray}
\frac{\mathrm{d} y(t)}{\mathrm{d} t} & = & f(y(t),t) \label{eq:ode-original}
\end{eqnarray}
\noindent where $f(y(t),t)$ defines a time-dependent vector field if we know its value at all points of $y$ and all instants of time $t$. Eq. (\ref{eq:ode-original}) covers a broad range of problems, in that the change of a variable is determined by its current value and a time variable $t$. This formulation also works with Pre-norm Transformer blocks. For notational simplicity, we re-define $G(\textrm{LN}(y_t), \theta_t)$ as a new function $F(y_t, \theta_t)$:
\begin{eqnarray}
F(y_t, \theta_t) & = & G(\textrm{LN}(y_t), \theta_t)) \label{eq:g-f}
\end{eqnarray}
\noindent We then relax $y_t$ and $\theta_t$ to continuous functions $y(t)$ and $\theta(t)$, and rewrite Eq. (\ref{eq:prenorm}) to be:
\begin{eqnarray}
y(t+\Delta t) & = & y(t) + \Delta t \cdot F(y(t), \theta(t)) \label{eq:prenorm-delta}
\end{eqnarray}
\noindent where $\Delta t$ is the change of $t$, and is general called \textit{step size}. Obviously, we have $\Delta t = 1$ in Transformer. But we can adjust step size $\Delta t$ using a limit, and have
\begin{eqnarray}
\lim_{\Delta t \to 0} \frac{y(t+\Delta t) - y(t)}{\Delta t} & = & F(y(t), \theta(t)) \label{eq:prenorm-limit}
\end{eqnarray}
Given the fact that $\lim_{\Delta t \to 0} \frac{y(t+\Delta t) - y(t)}{\Delta t} = \frac{\mathrm{d} y(t)}{\mathrm{d} t}$, Eq. (\ref{eq:prenorm-limit}) is an instance of Eq. (\ref{eq:ode-original}). The only difference lies in that we introduce $\theta(t)$ into the right-hand side of Eq. (\ref{eq:ode-original}). Then, we say that a Pre-norm Transformer block describes an ODE. It has been found that Eq. (\ref{eq:prenorm}) shares the same form as the Euler method of solving the ODE described in Eq. (\ref{eq:prenorm-limit}) \cite{haberruthotto2017}. This establishes a relationship between Transformer and ODEs, in that, given $F(\cdot,\cdot)$ and learned parameters $\{\theta_t\}$, the forward pass of a multi-block Transformer is a process of running the Euler method for several steps.

\section{The ODE Transformer}
\label{sec:ode-transformer}

In numerical methods of ODEs, we want to ensure the precise solutions to the ODEs in a minimum number of computation steps. But the Euler method is not ``precise'' because it is a first-order method, and naturally with local truncation errors. The global error might be larger if we run it for a number of times.\footnote{The global error is what we would ordinarily call the error: the difference between $y(t)$ and the true solution. The local error is the error introduced in a single step: the difference between $y(t)$ and the solution obtained by assuming that $y(t-1)$ is the true solution} This is obviously the case for Transformer, especially when the multi-layer neural network arises a higher risk of instability in solving the ODEs \cite{haberruthotto2017}.

\subsection{High-Order ODE Solvers}

Here we use the Runge-Kutta methods for a higher order solution to ODEs \cite{runge1895numerische,kutta1901beitrag,butcher1996history,ascher1998computer}. They are a classic family of iterative methods with different orders of precision.\footnote{A $p$-order numerical method means that the global truncation error is proportional to $p$ power of the step size.} More formally, the explicit Runge-Kutta methods of an $n$-step solution is defined to be:
\begin{eqnarray}
y_{t+1} & = & y_{t} + \sum_{i=1}^{n}\gamma_i F_i \label{eq:rk-yt1} \\
    F_1 & = & h f(y_{t},t) \label{eq:rk-y1}\\
    F_i & = & h f(y_{t} + \sum_{j=1}^{i-1} \beta_{ij}F_j, t + \alpha_i h) \label{eq:rk-fi}
\end{eqnarray}
\noindent where $h$ is the step size and could be simply 1 in most cases. $F_i$ is an intermediate approximation to the solution at step $t + \alpha_i h$. $\alpha$, $\beta$ and $\gamma$ are coefficients which can be determined by the Taylor series of $y_{t+1}$ \cite{butcher1963coefficients}. Eq. (\ref{eq:rk-fi}) describes a sequence of solution approximations $\{F_1,...,F_n\}$ over $n$ steps $\{t+\alpha_1 h,...,t+\alpha_n h\}$. These approximations are then interpolated to form the final solution, as in Eq. (\ref{eq:rk-yt1}).

The Runge-Kutta methods are straightforwardly applicable to the design of a Transformer block. All we need is to replace the function $f$ (see Eq. (\ref{eq:rk-fi})) with the function $F$ (see Eq. (\ref{eq:g-f})). The advantage is that the function $F$ is re-used in a block. Also, the model parameter $\theta_t$ can be shared within the block.\footnote{Although we could distinguish the parameters at different steps in a block, we found that it did not help and made the model difficult to learn.} In this way, one can omit $t+\alpha_i h$ in Eq. (\ref{eq:rk-fi}), and compute $F_i$ by
\begin{eqnarray}
  F_i & = & F(y_{t} + \sum_{j=1}^{i-1} \beta_{ij}F_j, \theta_t)
  \end{eqnarray}
\noindent This makes the system more parameter-efficient. As would be shown in our experiments, the high-order Runge-Kutta methods can learn strong NMT systems with significantly smaller models.

The Runge-Kutta methods are general. For example, the Euler method is a first-order instance of them. For a second-order Runge-Kutta (RK2) block, we have
\begin{eqnarray}
  y_{t+1} & = & y_{t} + \frac{1}{2} (F_1 + F_2) \label{eq:rk2} \\
      F_1 & = & F(y_t, \theta_t) \\
      F_2 & = & F(y_t + F_1, \theta_t)
  \end{eqnarray}
\noindent This is also known as the improved Euler method. Likewise, we can define a fourth-order Runge-Kutta (RK4) block to be:
\begin{eqnarray}
  y_{t+1} & = & y_{t} + \nonumber \\
          &   & \frac{1}{6} (F_1 + 2F_2+ 2F_3+ F_4) \label{eq:rk4} \\
      F_1 & = & F(y_t, \theta_t) \\
      F_2 & = & F(y_t + \frac{1}{2}F_1, \theta_t)\\
      F_3 & = & F(y_t + \frac{1}{2}F_2, \theta_t)\\
      F_4 & = & F(y_t + F_3, \theta_t)
  \end{eqnarray}

% ##########Figure 2############
\input{figure2}
% ##########Figure 2############

See Figure \ref{fig:architecture} for a comparison of different Runge-Kutta blocks. It should be noted that the method presented here can be interpreted from the perspective of representation refinement \cite{iclrGreff2017}. It provides a way for a function to update the function itself. For example, Universal Transformer refines the representation of the input sequence using the same function and the same parameters in a block-wise manner \cite{Dehghani2019universal}. Here we show that inner block refinements can be modeled with good theoretical support.

\subsection{Coefficient Learning}
\label{sec:coefficient}

In our preliminary experiments, the RK2 and RK4 methods yielded promising BLEU improvements when the model was shallow. But it was found that the improvements did not persist for deeper models. To figure out why this happened, let us review the Runge-Kutta methods from the angle of training. Take the RK2 method as an example. We rewrite Eq. (\ref{eq:rk2}) by substituting $F_1$ and $F_2$, as follow
\begin{eqnarray}
  y_{t+1} & = & y_{t} + \frac{1}{2}F(y_{t},\theta_t) + \nonumber \\
          &   & \frac{1}{2}F(y_{t}+F(y_{t},\theta_t),\theta_t) \label{eq:rk2-full}
  \end{eqnarray}
Let $\mathcal{E}$ be the loss of training, $L$ be the number blocks of the model, and $y_{L}$ be the model output. The gradient of $\mathcal{E}$ at $y_t$ is
\begin{eqnarray}
  \frac{\partial \mathcal{E}}{\partial y_{t}} & = & \frac{\partial \mathcal{E}}{\partial {y_{L}}} \cdot \frac{1}{2^{L-t}} \cdot \prod_{k=t}^{L-1} (1+g_{k}) \label{eq:rk2-gradient}
  \end{eqnarray}
\noindent where
\begin{eqnarray}
  { g_{k}} & = & \hspace{-0.5em} \Big( 1+\frac{\partial F(y_{k},\theta_k)}{\partial y_{k}} \Big) \cdot \nonumber \\
                                          &   & \hspace{-0.5em} \Big( 1+\frac{\partial F(y_{k}+F(y_{k},\theta_k),\theta_k)}{\partial y_{k}+F(y_{k},\theta_k)} \Big) \label{eq:g-gradient}
  \end{eqnarray}
\noindent Seen from Eq. (\ref{eq:rk2-gradient}), $\frac{\partial \mathcal{E}}{\partial y_{t}}$ is proportional to the factor $\frac{1}{2^{L-t}}$. This leads to a higher risk of gradient vanishing when $L$ is larger.

The problem somehow attributes to the small coefficients of $F_i$, that is, $\gamma_1 = \gamma_2 = \frac{1}{2}$. A natural idea is to empirically set $\gamma_i = 1$ to eliminate the product factor of less than 1 in gradient computation, although this is not theoretically grounded in standard Runge-Kutta methods. We rewrite Eq. (\ref{eq:rk2-full}) with the new coefficients, as follows
\begin{eqnarray}
  y_{t+1} & = & y_{t} + F(y_{t},\theta_t) + \nonumber \\
          &   & F(y_{t}+F(y_{t},\theta_t),\theta_t) \label{eq:rk2-full-new}
  \end{eqnarray}
Then, we have the gradient, like this
\begin{eqnarray}
  \frac{\partial \mathcal{E}}{\partial y_{t}} & = & \frac{\partial \mathcal{E}}{\partial y_{L}} \cdot \prod_{k=t}^{L-1} g_{k} \label{eq:rk2-gradient-new}
  \end{eqnarray}
This model is easy to optimize because $\frac{\partial \mathcal{E}}{\partial_{y_{L}}}$ can be passed to lower-level blocks with no scales. Note that, the methods here are instances of parameter sharing \cite{Dehghani2019universal,lan2020albert}. For example, in each ODE block, we use the same function $F$ with the same parameter $\theta_t$ for all intermediate steps. Setting $\gamma_i = 1$ is a further step towards this because $F_i$ is passed to the following computations with the same scale. Here we call it implicit parameter sharing.

Another way of scaling $F_i$ to further improve ODE functions is to learn the coefficients automatically on the training data. The simplest method is to initialize $\gamma_i = 1$ and independently optimize each scale. It helps the system learn the way of flowing $F_i$ in a block. Based on it, scaling $F_i$ by a weighted gate mechanism \cite{srivastava2015highway} empirically achieves the best performance (see Section \ref{sec:exp}). Take RK2-block as an instance, the concatenation of $F_1$ and $F_2$ is transformed to a scalar $(0,1)$ through a $\textrm{sigmoid}$ gate, then the block output $y_{t+1}$ is
\begin{eqnarray}
  y_{t+1} & = & y_{t} + g \cdot F_1 + (1 - g) \cdot F_2  \label{eq:learnable}\\
  g & = & \textrm{sigmoid}([F_1, F_2] \cdot W + b)
  \label{eq:rk2-learnable}
  \end{eqnarray}
\noindent where $[,]$ denotes the concatenation operation and $W, b$ are learnable parameters. We call it RK2-block (learnable $\gamma_i$), and the architecture is shown in Figure \ref{fig:architecture} (d). This kind of formulation offers a more flexible way to decide which part contributes more and is also easy to be optimized. Moreover, we also summarize the comparison of various scaling functions in Appendix \ref{sec:more_results}.

%----------------------------------------------
\begin{table*}[t]
  \small
  % \vspace{-1cm}
  \setlength{\tabcolsep}{2.5pt}
  \centering
  \begin{tabular}{lrrrrrrrrr}
  \midrule
  \multirow{2}{*}{\textbf{Model}} & \multirow{2}{*}{\textbf{Layers \ }} & \multicolumn{4}{c}{\textbf{WMT En-De}}  & \multicolumn{4}{c}{\textbf{WMT En-Fr}} \\
  \cmidrule(r){3-6} \cmidrule(r){7-10}
  &  & \bf \#Param & \bf Steps & \bf BLEU & \bf SBLEU & \bf \#Param & \bf Steps & \bf BLEU & \bf SBLEU\\
    \midrule
    Transformer              \cite{vaswani2017attention}             &6-6    &213M  &100K   &28.40   &-      &222M  &300K  &41.00  &-     \\
    % Scaling NMT              \cite{ott2018scaling}                &6-6    &210M  &100K   &29.30   &28.6   &222M  &100K  &43.20  &41.4  \\
    MacaronNet               \cite{lu2019understanding}              &6-6    &-     &-      &30.20   &-      &-     &-     &-      &-     \\
    Depth growing            \cite{wu-etal-2019-depth}               &8-8    &270M  &800K   &29.92   &-      &-     &-     &43.27  &-     \\
    Transformer-DLCL         \cite{wang-etal-2019-learning}          &30-6   &137M  &50K    &29.30   &28.6   &-     &-     &-      &-     \\
    % Depth-wise Scale         \citet{zhang-etal-2019-improving}     &20-20  &560M  &300K   &29.62   &29.0   &108M  &300K  &40.58  &-     \\
    Multiscale Collaborative \cite{wei-etal-2020-multiscale}         &18-6   &512M  &300K   &30.56   &-      &-     &-     &-      &-     \\
    ADMIN                    \cite{liu-etal-2020-understanding}      &60-12  &262M  &250K   &30.01   &29.5   &-     &250K  &43.80  &41.8  \\
    SDT                      \cite{li-etal-2020-shallow}             &48-6   &192M  &50K    &30.21   &29.0   &198M  &100K  &43.28  &41.5  \\
    BERT-fused model         \cite{zhu2020incorporating}             &6-6    &-     &-      &30.75   &-      &-     &-     &43.78  &-     \\

    \midrule
    \multicolumn{10}{c}{\bf Base and Deep Models} \\
    \midrule
    Residual-block                              &6-6    &61M   &50K    &27.89       &26.8      &69M   &100K   &41.05       &39.1     \\
    % Weight-sharing                              &6-6    &61M   &50K    &28.51       &27.4      &69M   &100K   &41.65       &39.7     \\
    RK2-block                                   &6-6    &61M   &50K    &28.67       &27.5      &69M   &100K   &42.08       &40.1     \\
    RK2-block (learnable $\gamma_i$)            &6-6    &61M   &50K    &28.89       &27.7      &69M   &100K   &42.31       &40.3     \\
    RK4-block                                   &6-6    &61M   &50K    &29.03       &27.9      &69M   &100K   &42.56       &40.6     \\
    Residual-block                              &24-6   &118M  &50K    &29.43       &28.3      &123M  &100K   &42.67       &40.6     \\
    RK2-block                                   &24-6   &118M  &50K    &29.85       &28.7      &123M  &100K   &43.04       &41.1     \\
    RK2-block (learnable $\gamma_i$)            &24-6   &118M  &50K    &\bf 30.29   &\bf 29.2  &123M  &100K   &\bf 43.48    &\bf 41.5 \\
    RK4-block                                   &24-6   &118M  &50K    &29.80       &28.8      &123M  &100K   &43.28       &41.3     \\
    \midrule
    \multicolumn{10}{c}{\bf Wide Models} \\
    \midrule
    Residual-block-Big                          &6-6    &211M  &100K    &29.21       &28.1      &221M  &100K  &42.89       &40.9     \\
    RK2-block                                   &6-6    &211M  &100K    &30.11       &29.0      &221M  &100K  &43.34       &41.3     \\
    RK2-block (learnable $\gamma_i$)            &6-6    &211M  &100K    &30.53       &29.4      &221M  &100K  &43.59       &41.6     \\
    RK4-block                                   &6-6    &211M  &100K    &30.39       &29.3      &221M  &100K  &43.55       &41.6     \\
    Residual-block-Big                          &12-6   &286M  &100K    &29.91       &28.9      &297M  &100K  &43.22       &41.2     \\
    RK2-block                                   &12-6   &286M  &100K    &30.58       &29.4      &297M  &100K  &43.88       &42.0 \\
    RK2-block (learnable $\gamma_i$)            &12-6   &286M  &100K    &\bf 30.77   &\bf 29.6  &297M  &100K  &\bf 44.11    &\bf 42.2 \\
    RK4-block                                   &12-6   &286M  &100K    & 30.55   &29.4  &297M  &100K  &43.81    &41.9 \\

  \midrule
\end{tabular}
\caption{Comparison with the state-of-the-arts on the WMT En-De and WMT En-Fr tasks. We both report the tokenized BLEU and SacreBLEU scores for comparison with previous work.}
\label{tab:main-results}
\end{table*}
%----------------------------------------------

\subsection{Efficiency Discussion}
\label{sec:discussion}
ODE Transformer is efficient to use.
As we only apply the ODE design schema to the encoder side, it only brings minor impacts on the inference speed due to the autoregressive decoding schema.
Another concern here is memory consumption.
ODE Transformer consumes more memory than the baseline in the same depth since we need to store the intermediate approximations in the forward pass.
But the additional consumption is less than that of the baseline who has the same computation cost, which is acceptable for most scenarios.
We give a quantitative analysis in Section~\ref{sec:analysis}.

\section{Experiments}
\label{sec:exp}

We evaluated the ODE Transformer on three sequence generation tasks: machine translation, abstractive summarization and grammar error correction. The datasets we used are elaborated in the following section, and more details of experimental setups could be found in Appendix \ref{sec:exp_setup} and \ref{sec:training_evaluation}.
\subsection{Datasets}

\paragraph{Machine Translation}
We report results on three WMT benchmarks. For the WMT'14 English-German (En-De) task, the training data consisted of approximately $4.5$M tokenized sentence pairs, as in \cite{vaswani2017attention}. All sentences were segmented into sequences of sub-word units \cite{sennrich-subword-neural} with $32$K merge operations using a shared vocabulary. We selected \textit{newstest2013} as the validation data and \textit{newstest2014} as the test data. For the WMT'14 English-French (En-Fr) task, we used the dataset provided within Fairseq, i.e., $36$M training sentence pairs from WMT'14. \textit{newstest2012+newstest2013} was the validation data and \textit{newstest2014} was the test data. For the WMT'16 English-Romanian (En-Ro) task, we replicated the setup of \cite{mehta2020delight}, which used $600$K/$2$K/$2$K sentence pairs for training, evaluation and inference, respectively.

\paragraph{Abstractive Summarization} We also tested the models' ability to process long sequences on the CNN-DailyMail summarization task \cite{nallapati-etal-2016-abstractive,hermann2015teaching}. The preprocessed method was the same as in \cite{ott2019fairseq}. We used a shared BPE with $30$K operations, resulting in a vocabulary of $32,580$ entries. The evaluation metric was F1-Rouge \cite{lin-2004-rouge} (Rouge-1, Rouge-2 and Rouge-L).

\paragraph{Grammar Error Correction} We used the following datasets as the training data, including National University of Singapore Corpus of Learner English (NUCLE) \cite{dahlmeier-etal-2013-building}, Lang-8 Corpus of Learner English (Lang-8) \cite{tajiri-etal-2012-tense}, FCE dataset \cite{yannakoudakis-etal-2011-new}, and Write \& Improve + LOCNESS Corpus \cite{bryant-etal-2019-bea}. We borrowed the setup from \citet{chollampatt18multilayer} and used the provided preprocessed script. The word-level dropout technique was also applied to prevent the overfitting problem.

%----------------------------------------------
\begin{table}[t]
  \setlength{\tabcolsep}{1.5pt}
  \small
  \centering
  \begin{tabular}{lrrr}
  \toprule
  \bf Model & \bf Params & \bf Epochs & \bf BLEU \\
  \midrule
  Transformer in \citet{mehta2020delight}        & 62M & 170 & 34.30 \\
  DeLight \cite{mehta2020delight}                & 53M & 170 & 34.70 \\
  Int Transformer$^{\dag}$\cite{lin2020towards}  & -   & -   & 32.60 \\
  Transformer (Our impl.)                        & 69M & 20  & 33.49 \\
  RK2-block (learnable $\gamma_i$)               & 69M & 20  & 34.94 \\
  RK2-block-Big (learnable $\gamma_i$)           & 226M& 20  & \textbf{35.28} \\
  \bottomrule
  \end{tabular}
  \caption{Results on the WMT En-Ro task. $\dag$ indicates the related information is not reported.}
  \label{tab:en-ro}
\end{table}
%----------------------------------------------

\paragraph{Language Modeling}
The truncation error analysis is conducted on the Penn Treebank \cite{mikolov2011empirical}, which is a widely-used language model dataset. It contains $88$K, $3,370$ and $3,761$ sentences for training, validation and test. The vocabulary size was $10$K. We set the layer depth of the language model to $1$ or $2$ to make a fair comparison. Assume the layer depth is $1$, then the loss between the block output and the ground-truth could be regarded as the truncation error. It alleviates the influence of the error accumulation across different layers.

\subsection{Experimental Results}

\paragraph{Results of En-De and En-Fr}
Table \ref{tab:main-results} compares ODE Transformer with several state-of-the-art systems. Both RK2-block and RK4-block outperform the baselines by a large margin with different model capacities. For example, RK2-block obtains a $+1.00$ BLEU improvement with the base configuration when the depth is $6$. RK4-block yields a gain of $0.17$ BLEU points on top of RK2-block. This observation empirically validates the conjecture that high-order ODE functions are more efficient. When we switch to deep models, our method is more parameter efficient. E.g., RK2-block is comparable with a strong 48-layer system \cite{li-etal-2020-shallow} with half of the encoder depth.
Similarly, wide models can also benefit from the enlarging layer depth \cite{wei-etal-2020-multiscale,li-etal-2020-shallow}. RK2-block achieves BLEU scores of $30.77$ and $44.11$ on the En-De and the En-Fr tasks, significantly surpassing the standard Big model by $1.32$ and $0.70$ BLEU points. This sets a new state-of-the-art on these tasks with fewer parameters.

%----------------------------------------------
\begin{table}[t]
  \setlength{\tabcolsep}{1.5pt}
  \small
  \centering
  \begin{tabular}{lrr}
  \toprule
  \bf Model & \bf Params &  \bf BLEU \\
  \midrule
  Transformer \cite{vaswani2017attention}           & 62M  & 27.30 \\
  Evolved Transformer \cite{so2019evolved}          & 46M  & 27.70 \\
  Lite Transformer$^{\dag}$ \cite{wu2020lite}       & -    & 26.50 \\
  DeLight \cite{mehta2020delight}                   & 37M  & 27.60 \\
  RK2-block (learnable $\gamma_i$, H=256, L=28)     & 37M  & \textbf{28.24} \\
  RK2-block (learnable $\gamma_i$, H=256, L=18)     & 29M  & 27.84 \\
  \bottomrule
  \end{tabular}
\caption{The comparison of model efficiency on the WMT En-De task.}
\label{tab:en-de}
\end{table}
%----------------------------------------------

\paragraph{Results of En-Ro}
Table \ref{tab:en-ro} exhibits model parameters, total training steps and BLEU scores of several strong systems on the En-Ro task. Again, ODE Transformer outperforms these baselines. As stated in \cite{mehta2020delight}, they trained the model up to $170$ epochs and obtained a BLEU score of $34.70$ through the \texttt{DeLight} model. However, the observation here is quite different. The validation PPL begins to increase after $20$ epochs. Thus, our baseline is slightly inferior to theirs, but matches the result reported in \citet{lin2020towards}. ODE blocks achieve even better performance with \texttt{DeLight} within much less training cost. For a bigger model (line 6), it obtains a BLEU score of $35.28$.
% which is a new state-of-the-art on the En-Ro task.

\paragraph{Parameter Efficiency}
Table \ref{tab:en-de} summaries the results of several efficient Transformer variants, including Lite Transformer \cite{wu2020lite}, DeLight \cite{mehta2020delight} and a light version of the Evolved Transformer \cite{so2019evolved}. As expected, ODE Transformer is promising for smaller models. It is comparable in BLEU with \texttt{DeLight} but having $9$M fewer parameters. Under the same model capacity, it outperforms \texttt{DeLight} by $0.64$ BLEU points. It may offer a new choice for deploying NMT systems on edge devices.
% These results demonstrate that the proposed method is orthogonal to the model capacity.

\paragraph{Results of Summarization and Correction}
We also evaluated the ODE Transformer on another two sequence generation tasks. Table \ref{tab:summarization} shows that both RK2-block and RK4-block outperform the baselines by a margin. Similarly, RK4-block is superior to RK2-block when the model is shallow. More results and case studies could be found in Appendix \ref{sec:more_results}.
% Unlike \citet{liu2020understanding}'s work, we only applied word dropout rather than word dropout + SwitchOut \cite{wang-etal-2018-switchout}. Larger improvements would be expected if we use their setup.

\section{Analysis}
\label{sec:analysis}
Here we investigate some interesting issues. For simplicity, we call RK2-block with coefficients initialized by 1 as RK2-block-v1, and learnable coefficients (Eq. (\ref{eq:learnable}) ) as RK2-block-v2.

\paragraph{Quantization of the Truncation Error}
In fact, we cannot obtain the ``true'' solution of each block output in NMT, because we mainly experimented on the encoder side. Instead, we tested our system on the language modeling task, where the perplexity between the single-layer model output and the ground truth could be regarded as the truncation error with no error propagations.
Table \ref{tab:truncate_error} shows the perplexities on the Penn Treebank dataset \cite{mikolov2011empirical}. All ODE Transformer variants reduce the errors significantly. RK4-order achieves the lowest PPL on both settings. In addition, RK2-block can even obtain a lower PPL than a 2-layer residual-block. The observation here again verifies larger ODE blocks behave superior to the standard residual block.

%----------------------------------------------
\begin{table}[t]
  \setlength{\tabcolsep}{1.5pt}
  \small
  \centering
    \begin{tabular}{lcccccc}
  \toprule
  \multirow{2}{*}{\textbf{Model}} & \multicolumn{3}{c}{\textbf{Summarization}}  & \multicolumn{3}{c}{\textbf{Correction}} \\
  \cmidrule(r){2-4} \cmidrule(r){5-7}
  &  \bf RG-1 & \bf RG-2 & \bf RG-L & \bf Prec. & \bf Recall & \bf $\textrm{F}_{0.5}$ \\
  \midrule
  \citet{liu2020understanding}    & 41.00  & 18.30    & 37.90   &66.80  &35.00 &56.60 \\
  Residual-block                             & 40.47  & 17.73    & 37.29   &67.97  &32.17 &55.61 \\
  RK2-block                                  & 41.58  & 18.57    & 38.41   &\bf 68.21  &35.30 &57.49 \\
  RK4-block                                  & \bf 41.83  & \bf 18.84    & \bf 38.68   & 66.20  &\bf 38.13 &\bf 57.71 \\
  \bottomrule

\end{tabular}
  \caption{Results of ODE Transformer on the summarization and correction tasks.}
  \label{tab:summarization}
\end{table}
%----------------------------------------------

\paragraph{Inference Speed and Memory Consumption}
Table \ref{tab:inference_memory} shows the comparison of inference speed and memory consumption discussed in Section \ref{sec:discussion}. Experimental results demonstrate the proposed ODE design schema results in acceptable inference speeds. And it is also memory-friendly through the memory comparison between the baseline and the RK variants in both base and big configurations.

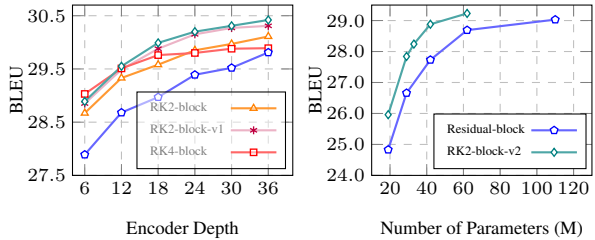
\begin{figure}[!t]
    \centering
    \begin{tikzpicture}
      \scriptsize{
        \begin{axis}[
       at={(0,0)},
        ymajorgrids,
        xmajorgrids,
        grid style=dashed,
        width=.28\textwidth,
        height=.24\textwidth,
        legend style={at={(0.46,0.13)}, anchor=south west},
        xlabel={\scriptsize{Encoder Depth}},
        ylabel={\scriptsize{BLEU}},
        ylabel style={yshift=-2em},xlabel style={yshift=0.0em},
        yticklabel style={/pgf/number format/precision=1,/pgf/number format/fixed zerofill},
        ymin=27.5,ymax=30.7, ytick={27.50,28.50,29.50,30.50},
        xmin=4,xmax=40,xtick={6,12,18,24,30,36},
        legend style={yshift=-6pt,xshift=-2em, legend plot pos=right,font={\tiny},cells={anchor=west},fill opacity=0.5, legend columns=1}
        ]
  
       % using "mark options" do more changes for marks
          
        \addplot[orange!80,mark=triangle*,,mark size=1.5pt,thick,mark options={fill=white,draw=orange,line width=0.5pt}] coordinates {(6,28.67) (12,29.33) (18,29.58) (24,29.85) (30,29.97) (36,30.11)};
        \addlegendentry{\scalebox{.8}{RK2-block}}
  
        \addplot[purple!30,mark=asterisk,mark size=1.5pt,thick,mark options={fill=white,draw=purple,line width=0.5pt}] coordinates {(6,28.85) (12,29.49) (18,29.88) (24,30.15) (30,30.27) (36,30.31) };
        \addlegendentry{\scalebox{.8}{RK2-block-v1}}
  
        \addplot[red!60,mark=square*,mark size=1.2pt,thick,mark options={fill=white,draw=red,line width=0.5pt}] coordinates {(6,29.03) (12,29.51) (18,29.76) (24,29.80) (30,29.88) (36,29.89)
        };
        \addlegendentry{\scalebox{.8}{RK4-block}}
  
        \addplot[blue!60,mark=pentagon*,mark size=1.5pt,thick,mark options={fill=white,draw=blue,line width=0.5pt}] coordinates {(6,27.89) (12,28.68) (18,28.97) (24,29.39) (30,29.52) (36,29.81) };
        % \addlegendentry{\scalebox{.8}{Base}}
  
        \addplot[teal!70,mark=diamond*,mark size=1.5pt,thick,mark options={fill=white,draw=teal,line width=0.5pt}] coordinates {(6,28.89) (12,29.55) (18,29.99) (24,30.20) (30,30.31) (36,30.42) };
        % \addlegendentry{\scalebox{.8}{RK2-v2}}

        \end{axis}
       }
  
      \scriptsize{
      %  \pgfplotsset{compat=1.11}
        \begin{axis}[
       at={(16em,0)},
        ymajorgrids,
        xmajorgrids,
        grid style=dashed,
        width=.28\textwidth,
        height=.24\textwidth,
        legend style={at={(0.45,0.13)}, anchor=south west},
        xlabel={\scriptsize{Number of Parameters (M)}},
        ylabel={\scriptsize{BLEU}},
        ylabel style={yshift=-2em},xlabel style={yshift=0.0em},
        yticklabel style={/pgf/number format/precision=1,/pgf/number format/fixed zerofill},
        ymin=24,ymax=29.5, ytick={24.00, 25.00,26.00,27.00,28.00,29.00},
        xmin=10,xmax=130,xtick={20,40,60,80,100,120},
        legend style={yshift=-6pt,xshift=-2em, legend plot pos=right,font={\tiny},cells={anchor=west}}
        ]
        % \draw[|-|,line width=0.6pt, black!80, dashed, thick] (62,31.23) -- (110, 31.23);
         % using "mark options" do more changes for marks
        \addplot[blue!60,mark=pentagon*,mark size=1.5pt,thick,mark options={fill=white,draw=blue,line width=0.5pt}] coordinates { (19,24.83) (29,26.66) (42,27.73) (62,28.69) (110, 29.03)
        };
        \addlegendentry{\scalebox{.8}{Residual-block}}
  
        \addplot[teal!70,mark=diamond*,mark size=1.5pt,thick,mark options={fill=white,draw=teal,line width=0.5pt}] coordinates {(19, 25.96)(29,27.84) (33,28.24) (42,28.88) (62,29.23) };
        \addlegendentry{\scalebox{.8}{RK2-block-v2}}
  
        % \addplot[orange!80,mark=triangle*,,mark size=1.5pt,thick,mark options={fill=white,draw=orange,line width=0.5pt}] coordinates {(6,28.85) (12,29.33) (18,29.58) (24,30.15) (30,29.97) (36,30.11)};
        % \addlegendentry{\scalebox{.8}{RK2-block}}
  
        % \addplot[red!60,mark=square*,mark size=1.2pt,thick,mark options={fill=white,draw=red,line width=0.5pt}] coordinates {(6,29.05) (12,29.52) (18,29.97) (24,29.91) (30,30.1) (36,30.2) };
        % \addlegendentry{\scalebox{.8}{RK4-block}}
        \end{axis}
        % \draw[|-|,line width=0.6pt, gray!120, dashed, thick] (5.2,2.73) -- (6.39,2.73);
        % \draw[|-|,line width=0.6pt, gray!120, dashed, thick] (4.7,0.74) -- (4.7,2.17);
        % \draw[|-|,line width=0.6pt, gray!120] (5.2,1.90) -- (5.2,2.59);
  
       }
    \end{tikzpicture}
    \caption{The comparison of BLEU against different encoder depth and the number of model parameters.}\label{fig:encoder-detph}
  \end{figure}

\paragraph{BLEU against Encoder Depth}
Figure \ref{fig:encoder-detph} (left) depicts BLEU scores of several ODE Transformer variants and the baseline under different encoder depths. All ODE Transformer variants are significantly superior to the baseline when depth $\leq 24$. RK2-block-v2 almost achieves the best performance over all depths, especially when the model becomes deeper. Interestingly, Figure \ref{fig:encoder-detph} confirms again that ODE Transformer is parameter efficient, e.g., a 6-layer RK2-block is comparable with the 18-layer baseline system. Another finding here is RK4-block performs well on shallow models, but it is inferior to RK2-block when the depth is going deep. This is because original coefficients may cause the optimization problem in the backward propagation in deep models (see Section \ref{sec:coefficient}). Also, Figure \ref{fig:encoder-detph} (right) plots BLEU as a function of the model size when the hidden size is $256$. The RK2 method significantly surpasses the baseline using much fewer parameters.

%----------------------------------------------
\begin{table}[t]
  \setlength{\tabcolsep}{3.5pt}
  \small
  \centering
  \begin{tabular}{lcc}
    \toprule
    \bf Model  & \bf 1-Layer & \bf 2-Layer \\
    \midrule
    Residual-Block                   & 142.33 & 136.07 \\
    RK2-block                        & 131.80 & 123.12 \\
    RK2-block ($\gamma_i=1$)         & 132.67 & 123.90 \\
    RK2-block (learnable $\gamma_i$) & 128.48 & 121.02 \\
    RK4-block                        & \bf 126.89 & \bf 119.46 \\
    \bottomrule
    \end{tabular}
  \caption{Comparison of PPL on systems with different ODE blocks.}
  \label{tab:truncate_error}
\end{table}
%----------------------------------------------

%----------------------------------------------
\begin{table}[t]
  \setlength{\tabcolsep}{3.5pt}
  \small
  \centering
  \begin{tabular}{lrccrr}
    \toprule
    \multirow{2}{*}{\textbf{Model}} & \multirow{2}{*}{\textbf{Depth}} & \multicolumn{2}{c}{\textbf{Inference}}  & \multicolumn{2}{c}{\textbf{Memory}} \\
    \cmidrule(r){3-4} \cmidrule(r){5-6}
    & & \bf Base & \bf Big &  \bf Base & \bf Big \\
    \midrule
    Residual-Block     & 6   & 147.1 & 98.7 & 7.2  & 13.2 \\
    Residual-Block     & 12  & 141.3 & 94.5 & 10.9 & 18.7 \\
    Residual-Block     & 24  & 122.0 & 87.3 & 14.1 & 23.5 \\
    RK2-Block          & 6   & 141.6 & 93.9 & 8.5  & 15.1 \\
    RK4-Block          & 6   & 124.8 & 87.1 & 9.7  & 18.2 \\
    \bottomrule
    \end{tabular}
  \caption{Comparison of inference speed (sentences/s) and memory consumption (G).}
  \label{tab:inference_memory}
\end{table}
%----------------------------------------------

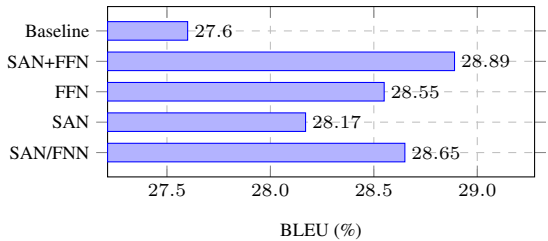
\begin{figure}[!t]
    \centering
    \begin{tikzpicture}
    \scriptsize{
    \begin{axis}[
      ymajorgrids,
      xmajorgrids,
      grid style=dashed,
      xbar,
      height=.24\textwidth,
      width=.45\textwidth,
      bar width=1em,
      xlabel={BLEU (\%)},
      symbolic y coords={{SAN/FNN}, {SAN}, {FFN}, {SAN+FFN}, {Baseline}},
      ytick=data,
      nodes near coords,
      nodes near coords align={horizontal},
      enlarge y limits=0.2,
      enlarge x limits=0.3,xticklabel style={/pgf/number format/fixed,/pgf/number format/fixed zerofill,/pgf/number format/precision=1},]
      \addplot[fill=blue!30, draw=blue] coordinates {(28.65,{SAN/FNN}) (28.17,{SAN}) (28.55,{FFN}) (28.89,{SAN+FFN}) (27.60,{Baseline})};
    \end{axis}
  }
    \end{tikzpicture}
    \caption{BLEU scores $[\%]$ of several $F(\cdot,\cdot)$ on the WMT En-De task.}
    \label{fig:components}
  \end{figure}

\paragraph{Ablation Study on Different $F(\cdot,\cdot)$}
As stated in Section \ref{sec:ode-transformer}, the $F(\cdot,\cdot)$ function can either be SAN, FFN or both of them (SAN+FFN). As shown in Figure \ref{fig:components}, high-order ODE works better with FFN than SAN. An explanation might be that the FFN component has more parameters than the SAN component.\footnote{There are $2 \cdot d_{\mathrm{model}}\cdot 4d_{\mathrm{model}}$ parameters in FFN and $d_{\mathrm{model}}\cdot 3d_{\mathrm{model}}$ + $d_{\mathrm{model}}\cdot d_{\mathrm{model}}$ in SAN.}
% There is a slight improvement when redesigning the SAN and the FFN simultaneously.
The model that treats FFN and SAN as a single ODE block behaves the best.

% ##########Figure 3############
%----------------------------------------------
  %----------------------------------------------
% ##########Figure 3############

% ##########Figure 4############
%----------------------------------------------
  %----------------------------------------------
% ##########Figure 4############

\paragraph{Training and Validation Perplexity}
Figure \ref{fig:loss} plots the training and validation PPL curves of RK blocks and the baseline enhanced by RPR \cite{shaw-etal-2018-self}. RK2-block obtains lower training and validation PPLs in both configurations (base and wide models).

\paragraph{Visualization of the Gradient Norm}
% To study the superiority of the proposed ODE Transformer,
We also collect the gradient information of several well-trained systems during training. Figure \ref{fig:gradient_norm} plots the gradient norm of RK2-block-v2, RK4-block and the standard residual-block (baseline). As we can see that Pre-Norm residual block is able to make the training stable \cite{wang-etal-2019-learning}. Both RK2-block-v2 and RK4-block provide richer signals due to the implicit parameter sharing among intermediate approximations. The two learning curves appear to be nearly the same, which is consistent with the results in Table \ref{tab:main-results}.

\paragraph{Comparison of Different ODE Design Schemas}
Then, we take a comprehensive analysis of several ODE design schemas. As stated in \citet{yiping2018beyond}'s work, several models in computer vision, such as LeapfrogNet \cite{he2019ode}, PolyNet \cite{zhang2017polynet} and MultistepNet \cite{yiping2018beyond}, can also be interpreted from the ODE perspective. The related ODE functions are summarized in Table \ref{tab:comparison}. We re-implemented these methods using the same codebase for fair comparisons. We conducted experiments following the base configuration on the En-De task.

% ##########Figure 5############
\definecolor{upurple}{RGB}{155,89,182}
\definecolor{ublue}{RGB}{52,152,219}
\definecolor{ured}{RGB}{231,76,60}
\definecolor{udark}{RGB}{77,153,77}
\definecolor{ugreen}{RGB}{46,204,113}
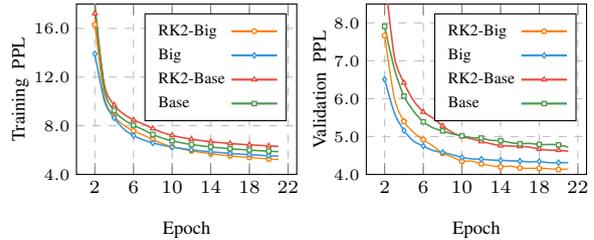
\begin{figure}[!t]
\centering
\begin{tikzpicture}
\scriptsize{
\begin{axis}[
at={(0,0)},
width=.28\textwidth, height=.24\textwidth ,
xtick={0,2,6,...,22},
ytick={4.00,8.00,...,18.00},
xlabel={Epoch},
grid style=dashed,
ylabel={Training\ \ PPL},
xlabel style={align=center,font=\scriptsize},
ylabel style={font=\scriptsize,yshift=-2em},
y tick style={opacity=0},
y tick label style={font=\tiny},
ymajorgrids=true,
xmajorgrids=true,
tick align=inside,
legend pos=outer north east,
yticklabel style={/pgf/number format/precision=1,/pgf/number format/fixed zerofill},
legend style={yshift=-0.5em,xshift=-8.5em,legend cell align=left,legend plot pos=right},
ymin=4.00,
ymax=18.00]

\addplot [sharp plot,orange,smooth,thick,line width=0.5pt,mark=pentagon*,mark size=1pt,thick,mark options={fill=white,draw=orange,line width=0.5pt}] coordinates {(0,0)};
\addplot [sharp plot,ublue,smooth,thick,line width=0.5pt,mark=diamond*,mark size=1pt,thick,mark options={fill=white,draw=ublue,line width=0.5pt}] coordinates {(0,0)};
\addplot [sharp plot,ured,smooth,thick,line width=0.5pt,mark=triangle*,,mark size=1pt,thick,mark options={fill=white,draw=ured,line width=0.5pt}] coordinates {(0,0)};
\addplot [sharp plot,udark,smooth,thick,line width=0.5pt,mark=square*,mark size=0.8pt,thick,mark options={fill=white,draw=udark,line width=0.5pt}] coordinates {(0,0)};

\addplot [sharp plot,orange,smooth,thick,line width=0.5pt] table [x=value,y=result,col sep=comma] {Figure/RK2-Big-train.csv};
\addplot [sharp plot,orange,mark=pentagon*,mark size=1pt,line width=0.5pt,only marks,mark options={fill=white,draw=orange,line width=0.5pt}] table [x=value,y=result,col sep=comma] {Figure/RK2-Big-train-new.csv};

\addplot [sharp plot,ublue,smooth,line width=0.6pt] table [x=value,y=result,col sep=comma] {Figure/RPR-Big-train.csv};
\addplot [sharp plot,ublue,mark=diamond*,mark size=1pt,smooth,line width=0.6pt,only marks,mark options={fill=white,draw=ublue,line width=0.5pt}] table [x=value,y=result,col sep=comma] {Figure/RPR-Big-train-new.csv};

\addplot [sharp plot,ured,smooth,thick,line width=0.6pt] table [x=value,y=result,col sep=comma] {Figure/RK2-Base-train.csv};
\addplot [sharp plot,ured,mark=triangle*,mark size=1pt,smooth,thick,only marks,mark options={fill=white,draw=ured,line width=0.5pt}] table [x=value,y=result,col sep=comma] {Figure/RK2-Base-train-new.csv};

\addplot [sharp plot,udark,smooth,line width=0.6pt] table [x=value,y=result,col sep=comma] {Figure/RPR-Base-train.csv};
\addplot [sharp plot,udark,mark=square*,mark size=0.8pt,smooth,only marks,mark options={fill=white,draw=udark,line width=0.5pt}] table [x=value,y=result,col sep=comma] {Figure/RPR-Base-train-new.csv};

\legend{\tiny{RK2-Big},\tiny{Big},\tiny{RK2-Base},\tiny{Base}},
\end{axis}
}
\vspace{6cm}
\scriptsize{
\begin{axis}[
at={(15.5em,0)},
width=.28\textwidth, height=.24\textwidth ,
xtick={0,2,6,...,22},
ytick={4.00,5.00,...,8.00},
xlabel={Epoch},
grid style=dashed,
ylabel={Validation\ \ PPL},
xlabel style={align=center,font=\scriptsize},
ylabel style={font=\scriptsize,yshift=-2.5em},
y tick style={opacity=0},
%x tick label style={font=\small},
y tick label style={font=\tiny},
ymajorgrids=true,
xmajorgrids=true,
tick align=inside,
legend pos=outer north east,
legend style={yshift=-0.5em,xshift=-8.5em,legend cell align=left,legend plot pos=right},
yticklabel style={/pgf/number format/precision=1,/pgf/number format/fixed zerofill},
ymin=4.00,
ymax=8.50]
%\addplot [sharp plot,orange,mark=otimes*,mark size=1pt,smooth,thick,line width=0.5pt] table [x=value,y=result,col sep=comma] {Figure/RK2-Big-valid.csv};
%\addplot [sharp plot,ublue, mark=+,mark size=1pt,smooth,thick,line width=0.6pt] table [x=value,y=result,col sep=comma] {Figure/RPR-Big-valid.csv};
%\addplot [sharp plot,ured,mark=star,mark size=1pt,smooth,dotted,thick] table [x=value,y=result,col sep=comma] {Figure/RK2-Base-valid.csv};
%\addplot [sharp plot,udark,mark=triangle*,mark size=1pt,smooth] table [x=value,y=result,col sep=comma] {Figure/RPR-Base-valid.csv};
\addplot [sharp plot,orange,smooth,thick,line width=0.5pt,mark=pentagon*,mark size=1pt,thick,mark options={fill=white,draw=orange,line width=0.5pt}] coordinates {(0,0)};
\addplot [sharp plot,ublue,smooth,thick,line width=0.5pt,mark=diamond*,mark size=1pt,thick,mark options={fill=white,draw=ublue,line width=0.5pt}] coordinates {(0,0)};
\addplot [sharp plot,ured,smooth,thick,line width=0.5pt,mark=triangle*,,mark size=1pt,thick,mark options={fill=white,draw=ured,line width=0.5pt}] coordinates {(0,0)};
\addplot [sharp plot,udark,smooth,thick,line width=0.5pt,mark=square*,mark size=0.8pt,thick,mark options={fill=white,draw=udark,line width=0.5pt}] coordinates {(0,0)};

\addplot [sharp plot,orange,smooth,thick,line width=0.5pt] table [x=value,y=result,col sep=comma] {Figure/RK2-Big-valid.csv};
\addplot [sharp plot,orange,mark=pentagon*,mark size=1pt,line width=0.5pt,only marks,mark options={fill=white,draw=orange,line width=0.5pt}] table [x=value,y=result,col sep=comma] {Figure/RK2-Big-valid-new.csv};

\addplot [sharp plot,ublue,smooth,line width=0.6pt] table [x=value,y=result,col sep=comma] {Figure/RPR-Big-valid.csv};
\addplot [sharp plot,ublue,mark=diamond*,mark size=1pt,smooth,line width=0.6pt,only marks,mark options={fill=white,draw=ublue,line width=0.5pt}] table [x=value,y=result,col sep=comma] {Figure/RPR-Big-valid-new.csv};

\addplot [sharp plot,ured,smooth,thick,line width=0.6pt] table [x=value,y=result,col sep=comma] {Figure/RK2-Base-valid.csv};
\addplot [sharp plot,ured,mark=triangle*,mark size=1pt,smooth,thick,only marks,mark options={fill=white,draw=ured,line width=0.5pt}] table [x=value,y=result,col sep=comma] {Figure/RK2-Base-valid-new.csv};

\addplot [sharp plot,udark,smooth,line width=0.6pt] table [x=value,y=result,col sep=comma] {Figure/RPR-Base-valid.csv};
\addplot [sharp plot,udark,mark=square*,mark size=0.8pt,smooth,only marks,mark options={fill=white,draw=udark,line width=0.5pt}] table [x=value,y=result,col sep=comma] {Figure/RPR-Base-valid-new.csv};

\legend{\tiny{RK2-Big},\tiny{Big},\tiny{RK2-Base},\tiny{Base}},
\end{axis}
}
\end{tikzpicture}
\caption{The comparison of training and validation PPL on base and wide models.}
\label{fig:loss}
\end{figure}
%----------------------------------------------
% ##########Figure 5############

% ##########Figure 6############
%----------------------------------------------
\begin{figure}[!t]
    \centering
    \begin{tikzpicture}
      \scriptsize{
      \begin{axis}[
      width=.46\textwidth, height=.24\textwidth ,
      xlabel=Step (K),
      ylabel=Value,
      xmin=0, xmax=55,
      ymin=0, ymax=0.7,
      xtick={0,10,20,30,40,50},
      ytick={0.1,0.2,0.3,0.4,0.5,0.6},
      yticklabels={$0.1$,$0.2$,$0.3$,$0.4$,$0.5$,$0.6$},
      ymajorgrids=true,
      xmajorgrids=true,
      grid style=dashdotted,
      legend cell align=left,
      scaled ticks=false,
      xlabel style={align=center,font=\scriptsize},
      ylabel style={font=\scriptsize,yshift=-2em},
      y tick style={opacity=0},
      x tick label style={font=\tiny},
      y tick label style={font=\tiny},
      legend style={yshift=-0.2em,xshift=0em,legend cell align=left,legend plot pos=right},
      ]
      \addplot [sharp plot,red,mark size=1pt,thick,line width=0.5pt,mark size=0.2pt] table [x=Step,y=Value,col sep=comma] {Figure/smooth3-2_avg3.csv};
      \addplot [sharp plot,blue,mark size=1pt,thick,line width=0.5pt,mark size=0.2pt] table [x=Step,y=Value,col sep=comma] {Figure/smooth3-3_avg3.csv};
      \addplot [sharp plot,orange,mark size=1pt,thick,line width=0.5pt,smooth] table [x=Step,y=Value,col sep=comma] {Figure/3-1.csv};
      \legend{\tiny{Residual-block},\tiny{RK2-block-v2},\tiny{RK4-block}},
      \end{axis}
      }
      % \pgfmathprintnumber[fixed zerofill,precision=5]{12345}
      \end{tikzpicture}
    \caption{Visualization of the gradient norm of ODE Transformers compared with the baseline.}
     \label{fig:gradient_norm}
    \end{figure}
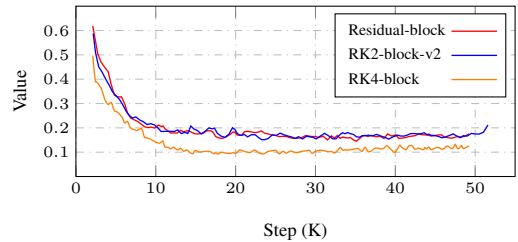
  %----------------------------------------------
  
% ##########Figure 6############

At the time $t$, Multistep Euler methods require previous states, e.g. $y_{t-1}$, to generate the current approximation, instead of iterative refinements based on the current-time state. So these methods are heavier than ODE Transformer. Note that DLCL \cite{wang-etal-2019-learning} can also be regarded as a multistep Euler method, which is more competitive in deep Transformer. But there is just a modest improvement upon the shallow baseline.
% However, our high-order design schema can achieve consistent BLEU improvement over different model depths.
Theoretically, the Backward Euler method is slightly better than the Forward Euler method in numerical analysis, but the improvement is marginal. Note that our ODE Transformer achieves consistent BLEU improvements over the aforementioned methods.
The reason is that such iterative refinements provide more efficient and effective parameter learning.

%----------------------------------------------
\begin{table*}[t]
  \setlength{\tabcolsep}{3.5pt}
  \renewcommand\arraystretch{1.2}
  \small
  \centering
  \begin{tabular}{llll}
  \toprule
  \bf Model  & \bf Information Flow & \bf Related ODEs &  \bf BLEU \\
  \midrule
  Leapfrog \cite{he2019ode}                & $y_{t+1} = y_{t-1}+2F(y_{t},\theta_t)$   & Multistep Euler         & 28.07 \\
  Multistep \cite{yiping2018beyond}        & $y_{t+1} = k_{n} \cdot y_{t}+(1-k_{n}) \cdot y_{t-1}+F(y_{t},\theta_t)$  & Multistep Euler         & 28.17 \\
  DLCL \cite{wang-etal-2019-learning}     & $y_{t+1} = y_{0} + \sum_{l=0}^{t}W_{l}{F(y_{l},\theta_l)}$              & Multistep Euler         & 27.78 \\
  PolyNet \cite{zhang2017polynet}         & $y_{t+1} = y_{t} + F(y_{t},\theta_t) +  F(F(y_{t},\theta_t),\theta_t)$                                     & Backward Euler          & 28.15 \\
  RK2-block                                   & $y_{t+1} = y_{t} + \frac{1}{2}F(y_{t},\theta_t) +  \frac{1}{2}F(y_{t}+F(y_{t},\theta_t),\theta_t)$         & Improved Euler          & 28.67 \\
  RK2-block ($\gamma_i=1$)                    & $y_{t+1} = y_{t} + F(y_{t},\theta_t) +  F(y_{t}+F(y_{t},\theta_t),\theta_t)$                               & RK 2nd-order     & 28.77 \\
  RK2-block (learnable $\gamma_i$)            & $y_{t+1} = y_{t} + \gamma_{1} \cdot F(y_{t},\theta_t)+ \gamma_{2} \cdot F(y_{t}+F(y_{t},\theta_t),\theta_t)$        & RK 2nd-order     & 28.86 \\
  RK4-block                                   & $y_{t+1} = y_{t} + \frac{1}{6}F_{1}+ \frac{2}{6}F_{2} + \frac{2}{6}F_{3} + \frac{1}{6}F_{4}$               & RK 4th-order     & 29.03 \\
  \bottomrule
  \end{tabular}
  \caption{Comparison of several ODE-inspired design schemas on the En-De task. We re-implement and apply these methods into Transformer. Note that $y_{n}$ denotes the model input of layer n. Due to the limited space, we use $F_{i}$ to denote the intermediate representation, where $i \in [1,4]$.}
  \label{tab:comparison}
\end{table*}
%----------------------------------------------

\section{Related Work}

\paragraph{Deep Transformer models}
Recently, deep Transformer has witnessed tremendous success in machine translation, especially on WMT news tasks \cite{li-etal-2019-niutrans,zhang-etal-2020-niutrans,zhou-etal-2021-niutrans,tran-etal-2021-facebook}. A straightforward way is to shorten the path from upper-level layers to lower-level layers thus to alleviate the gradient vanishing or exploding problems \cite{bapna-etal-2018-training,wang-etal-2019-learning,wu-etal-2019-depth,wei-etal-2020-multiscale}. For deeper models, the training cost is nonnegligible. To speed up the training, an alternative way is to train a shallow model first and progressively increase the model depth \cite{li-etal-2020-shallow,dong2020towards}. Apart from the model architecture improvements, another way of easing the optimization is to utilize carefully designed parameter initialization strategies \cite{zhang-etal-2019-improving,xu-etal-2020-lipschitz,Huang2020improving,liu-etal-2020-understanding}. With the model capacity going larger, one can use LayerDrop \cite{fan2019reducing} or Skipping Sublayers \cite{li2021learning} to prevent deep models from the overfitting problem.
Note that ODE Transformer is orthogonal to the aforementioned methods, and we will test it on these methods in future work.

\paragraph{Ordinary Differential Equations}
The relationship between ResNet and ODEs was first proposed by \citet{weinan2017proposal}. This shows a brand-new perspective on the design of effective deep architectures. Moreover, the success of Neural ODENet \cite{chen2018neural} has attracted researchers. Some insightful architectures \cite{zhang2017polynet,larsson2017fractalnet,yiping2018beyond,he2019ode,zhu2018convolutional,lu2019understanding,meila2021momentum} can also be interpreted from the ODE perspective. But, in NLP, it is still rare to see studies on designing models from the ODE perspective. \citet{zhang2021continuous} proposed continuous self-attention models using the same merit with neural ODE.
Perhaps the most relevant work with us is \citet{dutta2021redesigning}'s work. They redesigned the Transformer architecture from a multi-particle dynamic system view in terms of efficiency. Unlike them, we show that the stacked first-order ODE blocks may cause error accumulation, thus hindering the model performance. We address this issue by introducing high-order blocks, and demonstrate significant performance improvements on three sequence generation tasks, which is complementary to \citet{baier2020n}'s work.

\section{Conclusions}
This paper explores the relationship between Transformer and ODEs.
We propose ODE Transformer to help the model benefit from high-order ODE solutions.
Experimental results on the three representative sentence generations tasks (i.e., machine translation, abstractive summarization, and grammatical error correction) show the effectiveness and efficiency of ODE Transformer.
It achieves $30.77$ and $44.11$ BLEU scores on the WMT'14 En-De and En-Fr benchmarks, setting a new state-of-the-art result on the En-Fr. Note that our code is publicly available at \url{https://github.com/libeineu/ODE-Transformer}.

\section*{Acknowledgments}
This work was supported in part by the National Science Foundation of China (Nos. 61732005 and 61876035), the National Key R\&D Project of China (No. 2019QY1801), the China HTRD Center Project (No. 2020AAA0107904) and Yunnan Provincial Major Science and Technology Special Plan Projects (Nos. 201902D08001905 and 202103AA080015). The authors would like to thank anonymous reviewers for their valuable comments. And thank Yufan Jiang for his helpful advice to improve the paper.

\bibliography{acl_2022}
\bibliographystyle{acl_natbib}

\appendix

 %----------------------------------------------
 \begin{table*}[t]
  % \vspace{-1cm}
  \setlength{\tabcolsep}{3.5pt}
  \centering
  \begin{tabular}{lrrrrrrrrrcr}
  \midrule
  \multirow{2}{*}{\textbf{Dataset}} & \multirow{2}{*}{\textbf{Vocab}} & \multicolumn{3}{c}{\textbf{Dataset}}  & \multicolumn{5}{c}{\textbf{Training}} & \multicolumn{2}{c}{\textbf{Inference}} \\
  \cmidrule(r){3-5} \cmidrule(r){6-10} \cmidrule(r){11-12}
  &  & \bf Train & \bf Dev & \bf Test & \bf Lr  & \bf Warmup & \bf Batch & \bf Steps & \bf WD & \bf Beam & \bf LP\\
  \midrule
  WMT'14 En-De   & 34040   & 4.5M  & 3000  & 3003  & 0.002   &16000  &80K   &50K &$\times$   &4  &0.6 \\
  WMT'14 En-Fr   & 44424   & 35.7M & 26822 & 3003  & 0.002   &16000  &320K  &100K &$\times$   &4  &0.6 \\
  WMT'16 En-Ro   & 34976   & 602K  & 1999  & 1999  & 0.002   &8000   &80K   &17K &$\times$   &5  &1.3 \\
  CNN/DailyMail  & 32584   & 287K  & 13368  & 11490  & 0.002   &8000   &160K  &50K &$\times$ &4  &2.0 \\
  CONLL          & 33136   & 827K  & 5448  & 1312  & 0.0015   &4000   &160K  &15K &\checkmark &6  &0.6 \\
  \bottomrule

\end{tabular}
\caption{Statistics of the datasets and hyperparameters for three sequence generation tasks. For the dataset, we both report the vocabulary size, sentence numbers of training, validation and test sets. For the training, Lr denotes the peaking learning rate and Warmup denotes the warmup step of the Adam optimizer. WD denotes whether we applied word dropout. For the inference, Beam and LP denote the beam size and length penalty, respectively.}
  \label{tab:dataset}
\end{table*}
%----------------------------------------------

\section{Experimental Setups}
\label{sec:exp_setup}

Table \ref{tab:dataset} summarizes the details of our datasets. We both present the sentences and tokens of each task. For the En-De and En-Fr tasks, the datasets used in this work could be found in \texttt{Fairseq}.\footnote{\url{https://github.com/pytorch/fairseq/tree/master/examples/scaling_nmt}} For the En-Ro task, we used the preprocessed dataset provided by \texttt{DeLight}.\footnote{\url{https://github.com/sacmehta/delight/blob/master/readme_files/nmt/wmt16_en2ro.md}} Note that we only shared the target embedding and the softmax embedding instead of a shared vocabulary between the source side and the target side. The CNN/DailyMail dataset consists of CNN stories\footnote{\url{https://drive.google.com/uc?export=download&id=0BwmD_VLjROrfTHk4NFg2SndKcjQ}} and Daily emails.\footnote{\url{https://drive.google.com/uc?export=download&id=0BwmD_VLjROrfM1BxdkxVaTY2bWs}} For the grammar error correction task (GEC), we conducted experiments on the CONLL dataset.\footnote{\url{https://www.cl.cam.ac.uk/research/nl/bea2019st}}

\section{Training and Evaluation}
\label{sec:training_evaluation}

\paragraph{Training}
As suggested in \citet{li-etal-2020-shallow}'s work, we adopted relative positional representation (RPR) \cite{shaw-etal-2018-self} for stronger baselines. Dense connections among layers \cite{wang-etal-2019-learning} are also applied for stable learning since the model is optimized with FP16 training. 
All experiments were trained on $8$ GPUs with $4,096$ tokens on each GPU. For the En-De and the En-Fr tasks, we employed the gradient accumulation strategy with a step of $2$ and $8$, respectively. We used the Adam optimizer \cite{kingma2014adam} whose hyperparameters were set to $(0.9, 0.997)$. The hyperparameters including the learning rate, the warmup step and the total training steps of three tasks could be found in Table \ref{tab:dataset}. Note that we trained Base/Deep and Big models for $50$K and $100$K steps on the En-De task. We regarded merging SAN and FFN as the default ODE block. In addition, main results were the average of three times running with different random seeds, and we averaged the last 5/10 checkpoints for fair comparisons with previous work. The detail of Base/Deep/Wide configurations is as follows:

\begin{itemize}

\item Base/Deep Model. The hidden size of self-attention was $512$, and the dimension of the inner-layer in FFN was $2,048$. We used $8$ heads for attention. For training, we set all dropout to $0.1$ as default, including residual dropout, attention dropout, ReLU dropout. Label smoothing $\epsilon_{ls}=0.1$ was applied to enhance the generation ability of the model. For deep models, we only enlarged the encoder depth considering the inference speed.

\item Wide (or Big) Model. We used the same architecture as Transformer-Base but with a larger hidden layer size $1,024$, more attention heads ($16$), and a larger feed forward inner-layer ($4,096$ dimensions). The residual dropout was set to $0.3$ for the En-De task and $0.1$ for the En-Fr task.
\end{itemize}

For the language modeling task, the hidden size was $512$, and the filter size of the FFN was $2,048$. We set all the dropout rates as $0.1$, including the residual dropout, attention dropout and ReLU dropout. Each model was trained up to 20 epochs, and most models achieved the lowest PPL on the validation set when the epoch is 10. Then the validation PPL began to increase, though the training PPL is still declining. The warmup step was $2,000$ and the batch size was $4,096$. The max learning rate was set to $0.0007$.

\paragraph{Evaluation}
For machine translation, we measured performance in terms of BLEU. Both tokenized BLEU and SacreBLEU\footnote{BLEU+case.mixed+numrefs.1+smooth.exp+\\tok.13a+version.1.2.12} scores were reported on the En-De and En-Fr tasks. Also, we reported tokenized BLEU scores on the En-Ro task. In addition, we measured Rouge-1, Rouge-2, Rouge-L for CNN/DailyMail and precision, recall, $\textrm{F}_{0.5}$ for CONLL. The beam size and length penalty of each task are summarized in Table \ref{tab:dataset}.

% We plot the validation PPL curves of the baseline and RK blocks as shown in Figure \ref{fig:ppl}. RK blocks achieve significantly lower PPL than the baseline whenever the decoder depth is 1 or 2. On the other hand, RK2-block is also superior to 2-layer residual block within the same computation cost.

\section{Additional Results and Analyses}
\label{sec:more_results}

\paragraph{Comparison on the CNN/DailyMail Dataset}
We summarize the previous results on the CNN/DailyMail dataset (See Table \ref{tab:summarization-appendix}). The performance was evaluated by ROUGE-1, ROUGE-2 and ROUGE-L, respectively. Intuitively, high-order ODE functions can significantly improve on top of the Euler method as well as several strong existing models.\footnote{We only compared models without using pre-training.} Again, RK4-block beats the baseline and RK2-block by up to $1.36$ and $0.25$ scores in terms of ROUGE-1, respectively.

%----------------------------------------------
\begin{table*}[t]
  \setlength{\tabcolsep}{7.5pt}
  \centering
  \begin{tabular}{lccc}
  \toprule
  \bf Model & \bf ROUGE-1 & \bf ROUGE-2 & \bf ROUGE-L \\
  \midrule
  LEAD3                                    & 40.24  & 17.70    & 36.45   \\
  NEUSUM \cite{zhou-etal-2018-neural}      & 41.59  & 19.01    & 37.98   \\
  PGNet \cite{see-etal-2017-get}           & 39.53  & 17.28    & 36.38   \\
  Soft Fusion \cite{liu2020understanding}  & 41.00  & 18.30    & 37.90   \\
  Bottom-Up Summarization \cite{gehrmann-etal-2018-bottom} & 41.22  & 18.68    & 38.34   \\
  \midrule
  Residual-block                           & 40.47  & 17.73    & 37.29   \\
  RK2-block                                & 41.58  & 18.57    & 38.41   \\
  RK4-block                                & \bf 41.83  & \bf 18.84    & \bf 38.68  \\
  \bottomrule
\end{tabular}
\caption{ROUGE scores of various models on the CNN/DailyMail dataset.}
  \label{tab:summarization-appendix}
\end{table*}
%----------------------------------------------

%----------------------------------------------
\begin{table*}[t]
  \setlength{\tabcolsep}{7.5pt}
  \centering
  \begin{tabular}{lcccc}
  \toprule
  \bf Model  & \bf $\gamma_1$ & \bf $\gamma_2$ & \bf 6-layer  & \bf 24-layer \\
  \midrule
  weight sharing                     & 1        & 1              & 28.51   & 29.60  \\
  RK2-block                          & 1/2      & 1/2            & 28.67   & 29.85  \\
  RK2-block ($\gamma_i=1$)           &1         & 1              & 28.77   & 30.01  \\
  RK2-block (learnable $\gamma_i=1$) & scalar   & scalar         & 28.80   & 30.13  \\
  RK2-block (learnable $\gamma_i$)   & sigmoid  & sigmoid        & 28.74       & 30.06     \\
  RK2-block (learnable $\gamma_i$)   & sigmoid  & (1 - sigmoid)  & 28.86   & 30.29  \\
  RK2-block (learnable $\gamma_i$)   & tanh     & tanh           & 28.45   & 29.47  \\
  \bottomrule
\end{tabular}
\caption{Comparison of various scaling functions on the WMT14' En-De dataset.}
  \label{tab:scale_function}
\end{table*}
%----------------------------------------------

\paragraph{Comparison of Various Scaling Methods}
We have emphasized the importance of automatic coefficient learning in Section 3.2. The forward pass of RK2-block can be described as $y_{t+1} =  y_{t} + \gamma_1 \cdot F_1 + \gamma_2 \cdot F_2$, where $\gamma_1$ and $\gamma_2$ are coefficients which can be numerical suggested or learnable. Here we exhibit the comparison of various scaling methods on the WMT'14 En-De dataset, and the results are listed in Table \ref{tab:scale_function}. We can see that RK2-block (learnable $\gamma_i$) equips with a single sigmoid gate (line 5 in Table \ref{tab:scale_function}) yields best results on both shallow and deep configurations. The observation here reveals that appropriate scaling functions can further improve the RK2-block. Tanh activation even brings negative impacts on the performance, especially when the model is deep. A possible explanation is that Tanh produces a larger range ($[-1, 1]$) which is more difficult to optimize than the sigmoid function.

\begin{figure*}
    \begin{tikzpicture}
      \scriptsize{
      \begin{axis}[
    ymajorgrids=true,
        xmajorgrids=true,
        grid style=dashdotted,
        ybar,
        enlarge x limits=0.05,
        height=.5\textwidth,
        width=1\textwidth,
        bar width=0.5em,
        %xlabel={BLEU (\%)},
        %symbolic y coords={{SAN/FNN}, {SAN}, {FFN}, {SAN+FFN}, {Baseline}},
        symbolic x coords = {{R:NOUN:NUM},{R:PREP},{M:PUNCT},{U:DET},{R:VERB},{R:VERB:TENSE},{M:DET},{R:NOUN},{R:VERB:SVA},{R:DET},{U:OTHER},{R:VERB:FORM},{R:MORPH},{R:SPELL},{R:PRON},{M:OTHER},{M:PREP},{R:ORTH},{R:PUNCT},{U:PREP}},
        ymin=0,
        ymax=300,
        ytick = {0,50,100,150,200,250,300},
        xtick = data,
        %nodes near coords,
        %nodes near coords align={vertical},
        xticklabel style={/pgf/number format/fixed,/pgf/number format/fixed zerofill,/pgf/number format/precision=1, rotate=-90},]
        %\addplot[fill=blue!30, draw=blue] coordinates {(28.65,{SAN/FNN}) (28.17,{SAN}) (28.55,{FFN}) (28.89,{SAN+FFN}) (27.60,{Baseline})};
        \addplot[fill=blue!30, draw=blue!50] coordinates {({R:NOUN:NUM},285) ({R:PREP},238) ({M:PUNCT},234) ({U:DET},229) ({R:VERB},221) ({R:VERB:TENSE},187) ({M:DET},179) ({R:NOUN},175) ({R:VERB:SVA},165) ({R:DET},154) ({U:OTHER},141) ({R:VERB:FORM},130) ({R:MORPH},120) ({R:SPELL},112) ({R:PRON},105) ({M:OTHER},100) ({M:PREP},78) ({R:ORTH},76) ({R:PUNCT},67) ({U:PREP},60)};
        \addplot[fill=teal!50, draw=teal!70,xshift=-0.2em] coordinates {({R:NOUN:NUM},124) ({R:PREP},49) ({M:PUNCT},18) ({U:DET},108) ({R:VERB},13) ({R:VERB:TENSE},24) ({M:DET},83) ({R:NOUN},14) ({R:VERB:SVA},78) ({R:DET},24) ({U:OTHER},15) ({R:VERB:FORM},58) ({R:MORPH},33) ({R:SPELL},51) ({R:PRON},4) ({M:OTHER},1) ({M:PREP},30) ({R:ORTH},26) ({R:PUNCT},6) ({U:PREP},38)};
         \addplot[fill=red!40, draw=red!50,xshift=-0.4em] coordinates {({R:NOUN:NUM},148) ({R:PREP},95) ({M:PUNCT},47) ({U:DET},167) ({R:VERB},26) ({R:VERB:TENSE},50) ({M:DET},124) ({R:NOUN},26) ({R:VERB:SVA},99) ({R:DET},66) ({U:OTHER},27) ({R:VERB:FORM},69) ({R:MORPH},45) ({R:SPELL},54) ({R:PRON},19) ({M:OTHER},7) ({M:PREP},39) ({R:ORTH},29) ({R:PUNCT},21) ({U:PREP},52)};

    \node [minimum size=0.1em, fill=blue!30, draw=blue!50] (n1) at (9.4cm,5cm){};
    \node [anchor=west,] (w1) at ([xshift=0.5em]n1.east){Reference};
    \node [anchor=north,minimum size=0.1em, fill=teal!50, draw=teal!70] (n2) at ([yshift=-0.4em]n1.south){};
    \node [anchor=west,] (w2) at ([xshift=0.5em]n2.east){Residual-block};
    \node [anchor=north,minimum size=0.1em, fill=red!40, draw=red!50] (n3) at ([yshift=-0.4em]n2.south){};
    \node [anchor=west,] (w3) at ([xshift=0.5em]n3.east){RK4-block};
    \node [draw,minimum width=2.1cm,minimum height=1cm] at (10.25cm,4.7cm){};
      \end{axis}
    }
    \end{tikzpicture}
    \caption{Statistics of different error type information.}
    \label{fig:error-statistics}
    \end{figure*}
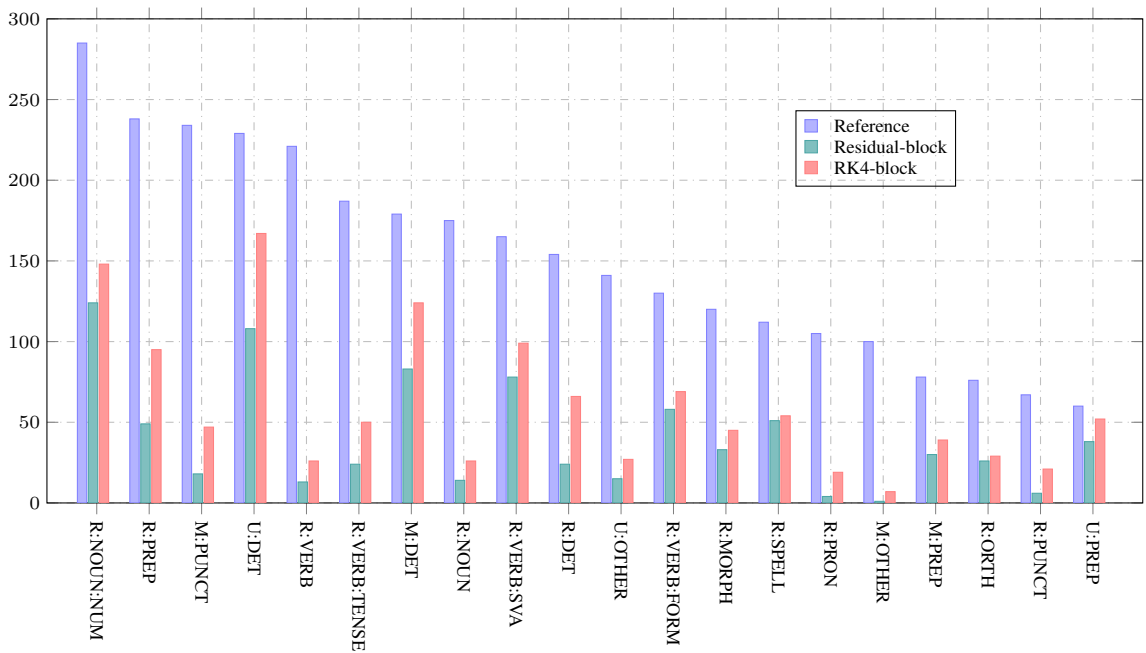

\paragraph{Case Study on the GEC Task}
Table \ref{tab:examples} summarizes several cases from the GEC task. Here, we make a comparison between the baseline and the RK4-block due to its superiority on the GEC task. We can clearly see that the proposed RK4-block delivers more accurate corrections compared with the baseline when handling subject-verb agreement (Case2), collocation (Case1, Case3), spelling (Case4) and other issues.
More specifically, Figure \ref{fig:error-statistics} illustrates the statistics of different error types annotated by ERRANT \cite{bryant-etal-2017-automatic}, a grammatical ERRor ANnotation Toolkit designed to automatically annotate parallel error correction data. For more details please refer to \citet{bryant-etal-2017-automatic}'s work. With the help of ERRANT, we can carry out a detailed error type analysis. As shown in Figure \ref{fig:error-statistics}, \textbf{\color{red!40}{RK4-block}} corrects the input in a more similar way with the \textbf{\color{blue!30}{reference}}, though there is still a large gap between them. Limited by the model ability, the \textbf{\color{teal!50}{baseline}} sometimes even cannot generate the right corrections, e.g. R:PUNCT and M:OTHER cases.

%----------------------------------------------
\begin{table*}[]
  \small
  \centering
  \begin{tabular}{p{0.8cm}|p{1.1cm}|p{12.2cm}}
    \toprule
    \multirow{4}*{Case1}&Source & What 's more , \textbf{various of} cultures can be shown to us through \textbf{social medias} .\\
    &Reference & What 's more , \textbf{\color{teal}{various}} cultures can be shown to us through \textbf{\color{teal}{social media}} .\\
    \cmidrule(r){2-3}
    &Baseline & What 's more ,\textbf{\color{teal}{ various}} cultures can be shown to us through \textbf{\color{red}{social medias}} .\\
    &RK4 & What 's more , \textbf{\color{teal}{various}} cultures can be shown to us through \textbf{\color{teal}{social media}} .\\
    \midrule
    \midrule
    \multirow{8}*{Case2} &Source & Social media sites such as Facebook \textbf{has allow} us to share our pictures or even chat online with our parents while we are overseas . \\
     &Reference & Social media sites such as Facebook \textbf{\color{teal}{have allowed}} us to share our pictures or even chat online with our parents while we are overseas .\\
     \cmidrule(r){2-3}
     &Baseline & Social media sites such as Facebook \textbf{\color{red}{allow}} us to share our pictures or even chat online with our parents while we are overseas .\\
     &RK4 & Social media sites such as Facebook \textbf{\color{teal}{have allowed}} us to share our pictures or even chat online with our parents while we are overseas .\\
     \midrule
     \midrule
    \multirow{4}*{Case3}&Source & \textbf{On one side} , it is \textbf{obvioualy} that many advantages have been brought to our lives . \\
     &Reference & \textbf{\color{teal}{On the one hand}} , it is \textbf{\color{teal}{obvious}} that many advantages have been brought to our lives .\\
     \cmidrule(r){2-3}
   &Baseline & \textbf{\color{red}{On one hand}} , it is \textbf{\color{teal}{obvious}} that many advantages have been brought to our lives .\\
     &RK4 & \textbf{\color{teal}{On the one hand}} , it is \textbf{\color{teal}{obvious}} that many advantages have been brought to our lives .\\
     \midrule
     \midrule
     \multirow{8}*{Case4}&Source & Other than that , I believe that the \textbf{stong} bond we have with our family is the biggest pillar of support to the carrier . \\
     &Reference & Other than that , I believe that the \textbf{\color{teal}{strong}} bond we have with our family is the biggest pillar of support to the carrier .\\
     \cmidrule(r){2-3}
     &Baseline & Other than that , I believe that the \textbf{\color{red}{stong}} bond we have with our family is the biggest pillar of support to the carrier .\\
     &RK4 & Other than that , I believe that the \textbf{\color{teal}{strong}} bond we have with our family is the biggest pillar of support to the carrier .\\
    \bottomrule
    \end{tabular}
    \caption{Several examples from the GEC task. Here, source and reference denote the model input and the correction result, respectively. \textbf{\color{teal}{Green}} words are good corrections, while \textbf{\color{red}{Red}} words are bad corrections.}
  \label{tab:examples}
    \end{table*}
  %----------------------------------------------

  %----------------------------------------------
  %----------------------------------------------

\section{Comparison with Related Work}
\label{sec:comparison_related}

As we aforementioned, the ODE design schema somehow shares a similar merit with the weight sharing, especially when the coefficients are set to 1. This is because we reuse the same function F to compute the intermediate approximation at each timestep, and it is also an effective way to apply the higher-order ODE into the Transformer architecture. Compared with weight sharing (line 1 in Table \ref{tab:scale_function}), ODE Transformer variants can deliver better performance within the same computation cost, demonstrating the effectiveness of ODE design schema.

Next, we make a detailed comparison between the proposed ODE Transformer and previous studies \cite{baier2020n,zhu2018convolutional,zhang2021continuous} to avoid the potential misunderstandings.

\paragraph{Compared with RKNet}
RKNet \cite{zhu2018convolutional} is mainly designed to improve the ResNet using implicit Runge-Kutta methods for vision tasks. There are some differences between ours and RKNet. (\romannumeral1) We mainly conduct experiments on sequence generation tasks, e.g. machine translation, abstract summarization, and grammar error correction tasks. They focused on the image classification task. (\romannumeral2) Except for the integration of ODE into the Transformer design schema, we also make an analysis on how to choose appropriate coefficients of intermediate approximations. And we bridge the relationship between the ODE design schema with the explicit weight sharing. (\romannumeral3) We also offer an automatic coefficient learning method for RK2-block which delivers the best performance in different configurations.

\paragraph{Compared with N-ODE}
As we discussed in the related work, our work is complementary to \citet{baier2020n}'s work. We empirically demonstrate the effectiveness of integrating ODE design schema into Transformer on several sequence generation tasks. This work may shed light on the design of effective Transformer architectures from the numerical perspective and provides stronger baselines to the literature.

\paragraph{Compared with CSAODE}
The differences between these two works are summarized below:  (\romannumeral1) As we emphasized above, the benchmarks we experimented on are quite different. They mainly validated the proposed CSAODE on text classification and QA tasks.  (\romannumeral2)  The proposed CSAODE \cite{zhang2021continuous} is an extension of neural ODE (cheng et al., 2018), where the motivation is quite different. They aim to effectively calculate the contiguous states of hidden features only via one-layer parameters and proposed a self-attention solver to fix the issue. While our motivation is to employ higher-order ODE solutions to reduce the truncation errors produced by each layer. On the other hand, CSAODE is still a single-layer model, and ours is a multi-layer sequence-to-sequence model. We also show the comparison of different components based on higher-order ODE solutions (See Figure \ref{fig:components}).  (\romannumeral3) The single-layer model is not strong enough to solve complicated tasks, e.g. machine translation. However, when stacking several layers, we need to re-consider the error accumulation among layers, that each layer is an individual ODE solver. How to mitigate the error accumulation is the main goal in this work, which is not discussed in their work.

\section{Derivations of the Equation}

Let $\mathcal{E}$ be the loss of training, $L$ be the number blocks of the model, and $y_{L}$ be the model output.
Here, we define

\begin{eqnarray}
  z_k  &=& y_k + F(y_k,\theta_k)
\end{eqnarray}

Then the information flow of the RK2 method can be described as follows:

\begin{eqnarray}
  y_{k+1} &=& y_k + \frac{1}{2}F(y_k,\theta_k) + \nonumber \\
  &    &\frac{1}{2}F(y_k+F(y_k,\theta_k),\theta_k)\nonumber \\
  & =  & y_k + \frac{1}{2}F(y_k,\theta_k) + \frac{1}{2}F(z_k,\theta_k) \label{eq:RK2-new}
\end{eqnarray}

\noindent where $\frac{\partial z_{k}}{\partial y_{k}}= 1 + \frac{\partial F(y_k,\theta_k)}{\partial y_k}$.
In this way, the detail derivation of Eq. (\ref{eq:RK2-new}) is as follows:

\begin{eqnarray}
  {\tiny \frac{\partial y_{k+1}}{\partial y_{k}}}
  &  =  &\frac{1}{2} \cdot \Big(1 + 1 +\frac{\partial F(y_k,\theta_k)}{\partial y_k}+ \nonumber \\
  &    & \frac{\partial F(z_k,\theta_k)}{\partial z_k} \cdot \Big(1+\frac{\partial F(y_k,\theta_k)}{\partial y_k}\Big)\Big)\nonumber \\
  & =  & \frac{1}{2} \cdot \Big(1 + \Big(1+\frac{\partial F(z_k,\theta_k)}{\partial z_k}\Big) \cdot \nonumber \\
  &    & \Big(1+\frac{\partial F(y_k,\theta_k)}{\partial y_k}\Big)\Big)\label{eq:gk-new}
  \end{eqnarray}

  % \begin{eqnarray}
  %   {\tiny \frac{\partial y_{k+1}}{\partial y_{k}}}
  %   & =  & 1 + \frac{1}{2}\frac{\partial F(y_k,\theta_k)}{\partial y_k} + \frac{1}{2}\frac{\partial F(z_k,\theta_k)}{\partial z_k} \cdot \frac{\partial z_k} {\partial y_k} \nonumber \\
  %   &  =  &\frac{1}{2} \cdot \Big(1 + 1 +\frac{\partial F(y_k,\theta_k)}{\partial y_k}+\frac{\partial F(z_k,\theta_k)}{\partial z_k} \cdot \nonumber \\
  %   &    & \Big(1+\frac{\partial F(y_k,\theta_k)}{\partial y_k}\Big)\Big)\nonumber \\
  %   & =  & \frac{1}{2} \cdot \Big(1 + \Big(1+\frac{\partial F(z_k,\theta_k)}{\partial z_k}\Big) \cdot \nonumber \\
  %   &    & \Big(1+\frac{\partial F(y_k,\theta_k)}{\partial y_k}\Big)\Big)\label{eq:gk-new}
  %   \end{eqnarray}

With the chain rule, the error $\mathcal{E}$ propagates from the top layer $y_L$ to layer $y_t$ by the following formula:
  \begin{eqnarray}
        \frac{\partial \mathcal{E}}{\partial y_t} = \frac{\partial \mathcal{E}}{\partial y_L} \cdot \frac{\partial y_{L}}{\partial y_{L-1}} \cdot \frac{\partial y_{L-1}}{\partial y_{L-2}} \cdots \frac{\partial y_{t+1}}{\partial y_{t}} \label{eq:chain}
  \end{eqnarray}

Here we have

\begin{eqnarray}
  {\small \hspace{-1em} g_{k}} & = & \hspace{-0.7em} \Big( 1+\frac{\partial F(y_{k},\theta_k)}{\partial y_{k}} \Big) \cdot \Big(1+\frac{\partial F(z_k,\theta_k)}{\partial z_k} \Big) \nonumber
  \end{eqnarray}

Then, put the Eq. (\ref{eq:chain}) into Eq. (\ref{eq:gk-new}), the gradient of $\mathcal{E}$ at $y_t$ is

  \begin{eqnarray}
  \frac{\partial \mathcal{E}}{\partial y_{t}} & = & \frac{\partial \mathcal{E}}{\partial_{y_{L}}} \cdot \frac{1}{2^{L-t}} \cdot \prod_{k=t}^{L-1} (1+g_{k})
  \end{eqnarray}

 Similarly, we can easily obtain the gradient of RK2 method where $\gamma_i=1$:

  \begin{eqnarray}
        \frac{\partial \mathcal{E}}{\partial y_t} &=& \frac{\partial \mathcal{E}}{\partial y_L} \cdot g_{L-1} \cdot g_{L-2} \cdots g_{t} \nonumber \\
    &=& \frac{\partial \mathcal{E}}{\partial y_L} \cdot \prod_{k=t}^{L-1}g_k
  \end{eqnarray}

\end{document}